%% file: main.tex
\documentclass{article}

\usepackage[preprint]{corl_2021}
\usepackage{microtype}
\usepackage{graphicx}

\usepackage{mathtools}
\usepackage{siunitx}
\usepackage[font=small,tableposition=top]{caption}
\usepackage{booktabs} % for professional tables
\usepackage{amssymb}
\usepackage{amsmath,amsthm}
\usepackage{mathtools}
\usepackage[usenames,dvipsnames]{xcolor}
\usepackage{subcaption}
\usepackage{algorithm}
\usepackage[noend]{algpseudocode}
\usepackage[utf8]{inputenc}
\usepackage[english]{babel}
\usepackage[inline]{enumitem}

\usepackage{dsfont}
\theoremstyle{definition}
\usepackage{hyperref}
\usepackage{cleveref}
\usepackage{changepage}
\usepackage{algorithmicx} %algpseudocode:layout for algorithmicx
\usepackage{titlesec}
\titlespacing\section{0pt}{0pt plus 2pt minus 2pt}{0pt plus 2pt minus 2pt}
\titlespacing\subsection{0pt}{3pt plus 4pt minus 2pt}{0pt plus 2pt minus 2pt}
\titlespacing\subsubsection{0pt}{3pt plus 4pt minus 2pt}{0pt plus 2pt minus 2pt}

\newcommand{\todo}[1]{}
\renewcommand{\todo}[1]{{\color{red} TODO: {#1}}}

\newcommand{\incmtt}[1]{{\fontfamily{cmtt}\selectfont{#1}}} % typewriter style
\definecolor{darkblue}{rgb}{0.0,0.0,0.7} % for hyper-links
\renewcommand\algorithmiccomment[1]{%
  \textcolor{darkblue}{\scriptsize \incmtt{\hfill\#\,{#1}}}%
}
\algnewcommand{\LineComment}[1]{\textcolor{darkblue}{\scriptsize{\incmtt{\#\ #1}}}}

\title{ SimNet: Enabling Robust Unknown Object Manipulation from Pure Synthetic Data via Stereo}
\author{
Thomas Kollar$^{*,1}$
Michael Laskey$^{*,1}$
Kevin Stone$^{*,1}$
Brijen Thananjeyan$^{*,1,2}$
Mark Tjersland$^{*,1}$
}

\makeatletter
\def\thanks#1{\protected@xdef\@thanks{\@thanks
        \protect\footnotetext{#1}}}
\makeatother
 \thanks{$^{1}$Toyota Research Institute. $^{2}$University of California, Berkeley.}
  \thanks{$^*$ Denotes equal contribution, $\alpha-\beta$ ordering.}
 \thanks{Contact: \texttt{\{first name.last name\} @ tri.global}}

\begin{document}
\input{0-defs.tex}

\maketitle
\vspace{-0.7cm}
\begin{abstract}
Robot manipulation of unknown objects in unstructured environments is a challenging problem due to the variety of shapes, materials, arrangements and lighting conditions.  Even with large-scale real-world data collection, robust perception and manipulation of transparent and reflective objects across various lighting conditions remains challenging.  To address these challenges we propose an approach to performing sim-to-real transfer of robotic perception.  The underlying model, SimNet, is trained as a single multi-headed neural network using simulated stereo data as input and simulated object segmentation masks, 3D oriented bounding boxes (OBBs), object keypoints and disparity as output.  A key component of SimNet is the incorporation of a learned stereo sub-network that predicts disparity.  SimNet is evaluated on 2D car detection, unknown object detection and deformable object keypoint detection and significantly outperforms a baseline that uses a structured light RGB-D sensor. By inferring grasp positions using the OBB and keypoint predictions, SimNet can be used to perform end-to-end manipulation of unknown objects in both "easy" and "hard" scenarios using our fleet of Toyota HSR robots in four home environments. In unknown object grasping experiments, the predictions from the baseline RGB-D network and SimNet enable successful grasps of most of the ``easy" objects. However, the RGB-D baseline only grasps 35\% of the ``hard" (e.g., transparent) objects, while SimNet grasps 95\%, suggesting that SimNet can enable robust manipulation of unknown objects, including transparent objects, in unknown environments. Additional visualizations and materials are located at \url{https://tinyurl.com/simnet-corl}.

% A network that uses the output of a structured RGB-D sensor is used as a baseline.  
%and when evaluated on challenging scenes that include transparent objects and in direct sunlight, SimNet is able to robustly predict OBBs and keypoints
\end{abstract}

% Two or three meaningful keywords should be added here
%===============================================================================

\input{1-introduction}

\input{2-related-work}

\input{3-cost-volume}
\input{4-panoptic-predictions}
\input{5-experiments}

\input{6-conclusion}

% The maximum paper length is 8 pages excluding references and acknowledgements, and 10 pages including references and acknowledgements

\clearpage
% The acknowledgments are automatically included only in the final version of the paper.
\footnotesize

%===============================================================================

% no \bibliographystyle is required, since the corl style is automatically used.
% \clearpage
\begin{small}
\bibliography{main}
\end{small}
\normalsize
%\newpage
%\begin{small}
% \bibliography{corl}  % .bib
% \end{small}
\input{7-appendix}
% \input{9-covid_one_pager.tex}
\end{document}

%% file: 0-defs.tex
\newcommand\smallO{
  \mathchoice
    {{\scriptstyle\mathcal{O}}}% \displaystyle
    {{\scriptstyle\mathcal{O}}}% \textstyle
    {{\scriptscriptstyle\mathcal{O}}}% \scriptstyle
    {\scalebox{.6}{$\scriptscriptstyle\mathcal{O}$}}%\scriptscriptstyle
  }

% \NewDocumentCommand{\evalat}{sO{\big}mm}{%
%   \IfBooleanTF{#1}
%   {\mleft. #3 \mright|_{#4}}
%   {#3#2|_{#4}}%
% }

\newcommand{\brijen}[1]{\textcolor{blue}{(#1 --Brijen)}}
\newcommand{\mike}[1]{\textcolor{red}{(#1 --Mike)}}
\newcommand{\tom}[1]{\textcolor{purple}{(#1 --Tom)}}
\newcommand{\kevin}[1]{\textcolor{orange}{(#1 --Kevin)}}
\newcommand{\TODO}[1]{\textcolor{red}{(TODO:#1)}}
\def \costvolume{Stereo Cost Volume Network}
\def \costvolumes{Stereo Cost Volume Networks}
\def \costabbr{SCVN}
\def \net{SimNet}

%% file: 1-introduction.tex
\section{Introduction}
To successfully deploy robots into diverse, unstructured environments such as the home, robots must have robust and general behaviors that that can adapt to variations in lighting, furniture, and objects.  As such, robots must be able to to perceive and manipulate objects that they have not seen before. This is made more challenging by the fact that many everyday objects are made of reflective or transparent materials and are located in places that have harsh lighting (e.g., Figure~\ref{fig:manipulation}, right side).

% Data.
To perform well in these settings, large-scale real-world data collection is often performed~\cite{gupta2018robot,sajjan2020clear,kalashnikov2018qt}, which is costly, time-consuming, has limited flexibility and sometimes contains missing data (as is the case with RGB-D sensing of transparent objects).  An alternative is to use simulation to automatically label a large dataset of images~\cite{mahler2019learning,sundaresan2020learning,sajjan2020clear,xie2020best,tobin2017domain,danielczuk2019segmenting,akkaya2019solving,ganapathi2020learning,hoque2020visuospatial,seita2019deepIL}. Although physics-based and photorealistic simulations can enable end-to-end learning methods, rendering can take significant computational resources and artists must generate high-quality artefacts~\cite{sajjan2020clear,roberts2020hypersim}.  

\begin{figure}[t!]
    \centering
    \includegraphics[width=5.5in]{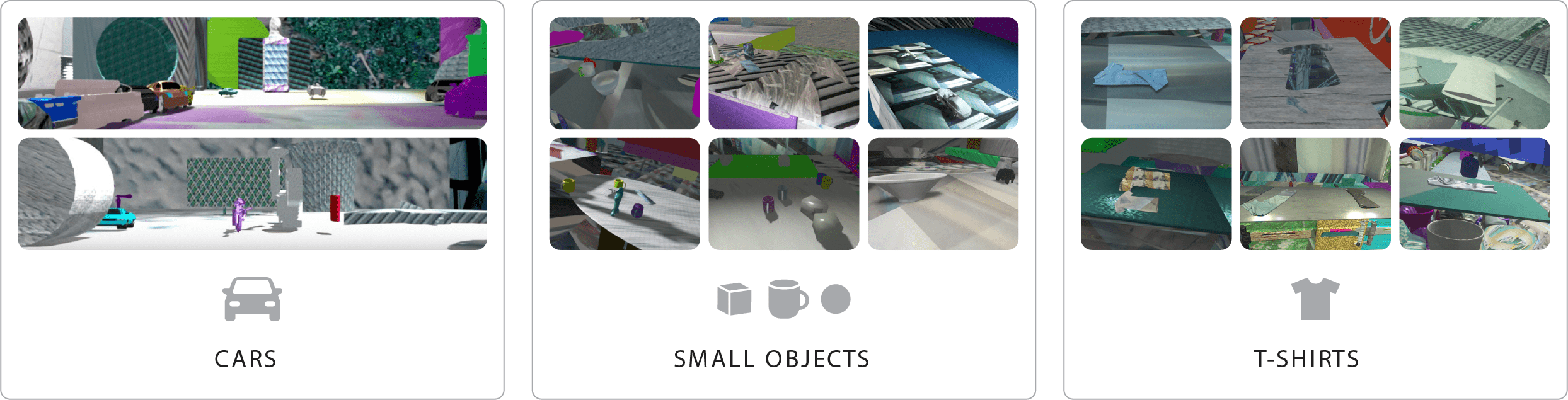}

\caption{\textbf{Simulation images:} Simulated data was generated for three domains: cars, small objects and t-shirts using a non-photorealistic simulator with domain-randomization.  Dataset generation is parallelized across machines and can be generated in an hour for \$60 (USD) cloud compute cost. Labels for OBBs, semantic segmentation masks, depth images, and keypoints can automatically be generated. The data is used to train \net{}. By learning to focus on geometry, sim2real transfer can be performed using only very low-quality scenes.
}
\label{fig:simulator}
\vspace{-4ex}
\end{figure}

This paper proposes an approach for sim-to-real transfer of robot perception and manipulation that does not require large-scale real-world datasets to be collected and does not require photorealism during simulation. The approach has three primary components. First, a non-photorealistic simulator that randomizes over lighting and textures is used to automatically produce large-scale domain-randomized data that includes stereo images~\cite{tobin2017domain,sadeghi2016cad2rl}, 3D oriented bounding boxes, object keypoints, and segmentation masks (Figure~\ref{fig:simulator}).  The second component and a core contribution \net{} (Figure~\ref{fig:simnet_diagram}) is a lightweight multi-headed neural network that is trained exclusively on simulated data and takes as input stereo pairs and predicts segmentation masks, 3D oriented bounding boxes (OBBs), keypoints and disparity as output. Key to our approach is the ability to predict disparity images using a differentiable cost volume to match features in a stereo image~\cite{mayer2016large,kendall2017end}. This network is end-to-end trained using a depth reconstruction auxiliary loss, which encourages the network to focus on disparity predictions and geometric information that are task-relevant. Unlike other approaches, which use active depth sensing~\cite{mahler2019learning,xie2020best,sajjan2020clear,sundaresan2020learning,danielczuk2019segmenting}, the incorporation of the learned differentiable stereo network into \net{} enables perception in the presence of non-Lambertian surfaces such as reflective or transparent materials and in challenging lighting conditions such as direct sunlight. \net{} can run at 20Hz on a Titan Xp GPU. The third component is a heuristic method for predicting grasp positions from the OBB and keypoint predictions. These grasp positions are combined with a classical planner to execute manipulation behaviors~\cite{grannen2020untangling,ganapathi2020learning,sundaresan2020learning,seita2018deep,sundaresan2019automated}.

% instead of producing the most accurate reconstructions of depth images, creating transferable geometric features that enable robust prediction for the "high-level" vision tasks
% thereby encouraging the use of geometric information and not simulation artifacts.

% Evaluation
\net{} is evaluated on three real-world domains: 2D car detection, unknown object grasping and deformable object keypoint prediction for t-shirt folding.  \net{} is trained on a large-scale synthetic dataset generated for each domain and evaluated on real-world validation sets that include optically-challenging objects, distractor objects, furniture and direct sunlight. The baseline network predicts the same outputs, but uses RGB-D input from a structured light sensor.  All three domains showed mAP gains with \net{} over the baseline. \net{} is also evaluated in an unknown object grasping experiment with both ``easy" and ``hard" objects using a fleet of Toyota HSR-C robots in four different homes.  The ``hard" objects are optically challenging and are reflective or transparent. Both the RGB-D network and \net{} are able to effectively generate grasps for the easy objects. However, the robot is only able grasp 35\% of the ``hard" objects with the RGB-D network, while it is able to grasp 95\% with \net{}. These results demonstrate that \net{} is able to grasp a wide variety of unknown objects, including transparent objects, in unstructured environments. \net{} is also evaluated on a t-shirt folding experiment across several homes and shirts.

%Despite the simplicity of the overall network and manipulation policies, \net{} performs significantly better than prior sim-to-real approaches that use active RGB-D sensing, achieving 95\% grasp success vs. 45\% on optically challenging objects such as glass and stainless steel.  Using the keypoint predictions, the ability to fold a t-shirt is also demonstrated.

% Contributions
This work makes five contributions: \textbf{\emph{(i)}} an efficient neural network, \net{}, that leverages learned stereo matching to enable robust multi-headed predictions of keypoints, oriented bounding boxes (OBBs), semantic segmentation masks, and depth images using only low-quality synthetic data, \textbf{\emph{(ii)}} the first network to enable single-shot prediction of 3D oriented bounding boxes of unknown objects \textbf{\emph{(iii)}} stereo vision experiments on simulated and real home scenes with annotations for oriented bounding boxes, segmentation masks, depth images, and keypoints, \textbf{\emph{(iv)}} experiments on the KITTI stereo vision benchmark dataset that suggest that the network transfers significantly better than standard domain randomization and naive stereo concatenation, and \textbf{\emph{(v)}} experiments in four homes that suggest that the network robustly transfers to real-world environments compared to active depth sensing and can be used to construct robot manipulation policies for grasping and t-shirt folding.  The model training code and validation datasets will be publicly released.

%% file: 2-related-work.tex
\section{Related Work}
% \tom{IMO, this needs to connect back to how we are doing manipulation.  How our approach to manipulation is similar.  For example, we should probably discuss other people who have used OBB-based appraoches for manipulation somewhere... not sure that is here right now.}
Our related work is divided into perception for manipulation, sim-to-real and learned stereo.

\begin{figure}[t!]
    \centering
    \includegraphics[width=5.in]{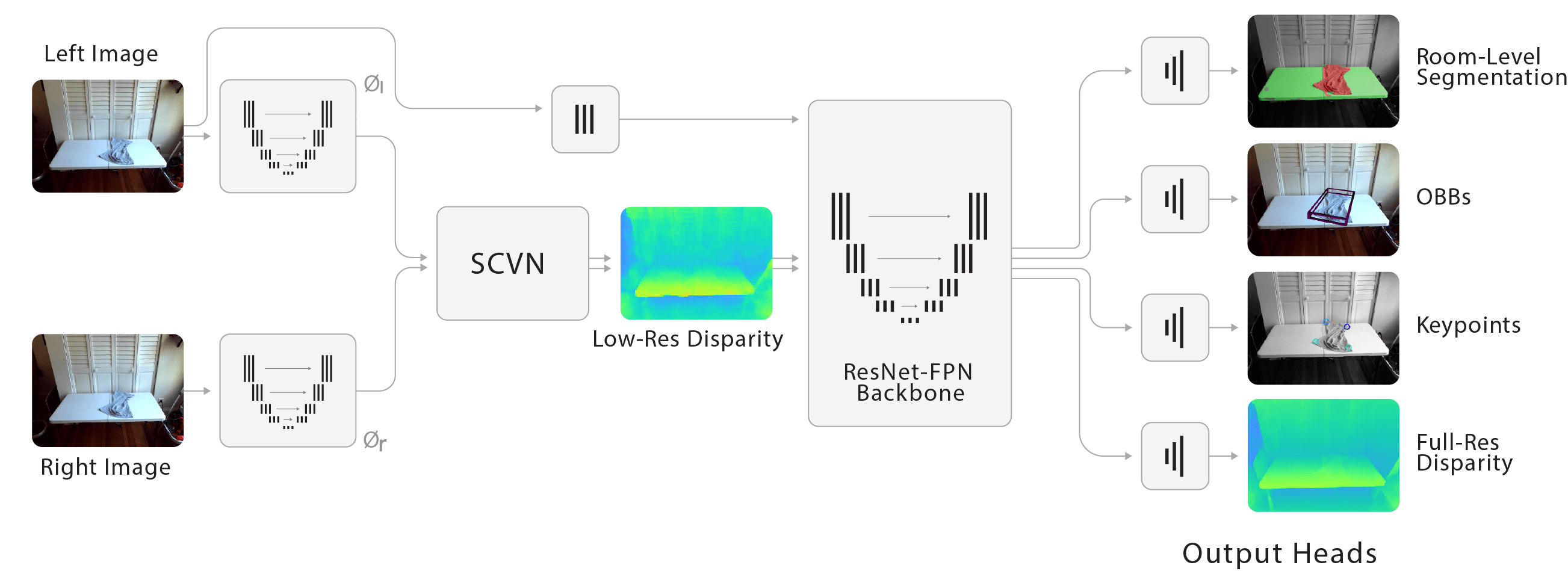}

\caption{\textbf{\net{}:}
% \tom{Add description to the figure.  This figure is the whole panoptic network.  SCVN is only a small piece of it.}}
% \brijen{extremely preliminary drafts of these figures. need to replace inputs with images/features not keypoint preds. replace gray boxes with something better} \mike{Couple of comments: 1) we should try to illustrate the internals of these blocks by drawing conv layers, you could draw a series of rectangles for example, 2) It would be helpful to illustrate the output heads for reference they are based on the semantic segmenation head illustrated in Fig 3. here https://arxiv.org/pdf/1901.02446.pdf. 3) With the SVCN, I am not sure if the top photo is conveying much, since the math is pretty clear in Sec. 3.2.  We might want to just have the bottom photo show to siamese net feature extractors feed into the SVCN block and a depth comes out}.
In \net{}, each stereo RGB image is fed into a feature extractor before being fed into \costabbr{}, which performs approximate stereo matching. The output of the \costabbr{} is a low resolution disparity image fed in with features from the left image to a ResNet-FPN backbone and output prediction heads. The output heads predict room-level segmentation, OBBs, keypoints, and full-resolution disparity images.}
\label{fig:simnet_diagram}
\vspace{-3ex}
\end{figure}
\vspace{-1ex}
\subsection{Parameterized Representations for Manipulation}
In contrast to end-to-end robot policy learning~\cite{kalashnikov2018qt,haarnoja2018soft,thananjeyan2021recovery,levine2016end},
% \tom{need to add citations here}
a popular approach for generating behaviors is to parameterize motions by the outputs of a perception network~\cite{mahler2019learning,sundaresan2020learning,grannen2020untangling,ganapathi2020learning,balasubramanian2012physical}. This decouples perception from planning and control, and enables perception systems to be trained in simulation without the need for accurate physical simulation.
One representation commonly used for grasping of rigid objects are oriented bounding boxes (OBBs)~\cite{balasubramanian2012physical}. These are useful for grasping common household objects, by aligning grasps along the principal components~\cite{balasubramanian2012physical}. Our work contributes a method to directly predict OBBs from a neural network. 
% \tom{IMO, you should probably lead with this (e.g., OBBs and the previous sentence for the whole related work section and then start contrasting).}
% \tom{As before, we need to be really clear what these perception networks are predicting and how they relate to our approach.} \mike{I would break this down into useful representation for deformable objects (i.e. keypoints) and rigid body (i.e. obb). It probably makese sense to have rigid body first}
% \brijen{I believe people also use keypoints for rigid objects, so it might not be a clear divide.}
A  representation that is used for deformable manipulation are keypoints or learned correspondences, which have seen significant success in tasks such as deformable manipulation~\cite{sundaresan2020learning,grannen2020untangling,ganapathi2020learning,maitin2010cloth,seita2018deep,seita2019deepIL} and grasping~\cite{manuelli2019kpam,florence2018dense}. \citet{ganapathi2020learning} and \citet{sundaresan2020learning} predict correspondences from domain-randomized depth or monocular RGB images and demonstrate impressive transfer to real fabrics and ropes in constrained lab environments. In contrast, we demonstrate robust transfer on diverse home scenes, and compare to the methods in these papers as baselines.
% \mike{We need to contrast our approach with theirs by talking about how it was limited to lab environment testing. We demonstrate when scaled up toe home scenes, our method is more robust, if the experiments hold} We show that \costabbr{}s more robustly transfer than these methods for sim-to-real transfer on several challenging domains \tom{shouldn't say what we are showing in the related work.  How is our approach different from them though?  Is it?}. 
% as an alternate representation, which simply capture the the statistics of a novel object's point cloud.
% \mike{Can you just cite that Balasumarian paper grasping paper here? Typically, OBBS have always been a common way to grasp unkown objects, where 6-DOF pose is for known objects. Since 6-DOF pose only makes since if you have a predefined model you are regressing a pose delta from} \brijen{What is the statement about novelty we can make about OBBs here, that we learn to predict them from stereo?}
% \brijen{In contrast to end-to-end, panoptic net, single shot prediction of these behavior parameters}

\subsection{Sim-to-Real Transfer}
Transferring perception models
% \tom{need to say what these models are doing... I assume not all of these are predicting the same things.  If they are, we should be explicit about what they are predicting.}
% \brijen{they typically are not all predicting the same thing. the same transfer techniques are ubiquitous and applied to a very wide array of robotics/perception tasks, each of which may have specialized requirements}
trained on simulated data to reality is an active area of research. One technique to transfer from simulation to reality is to train directly on a geometric representation of the world such as a point cloud or depth image~\cite{mahler2019learning,sundaresan2020learning}.  Prior work has used depth-based transfer techniques in a variety of scenarios, such as grasping unknown objects~\cite{mahler2019learning}, tying a rope~\cite{sundaresan2020learning}, and segmenting object instances in cluttered scenes~\cite{xie2020best,danielczuk2019segmenting}. Depth-based transfer is commonly performed with an active sensor using structured light. The applicability of these sensors is limited to regimes where their reflected infrared pattern is consistent enough that it can be pattern matched for 3-D reconstruction~\cite{grunnet2018projectors}. However, home settings contain non-Lambertian objects such as glassware and a large amount of natural light, which can create large sensing errors. Existing work studies combining depth sensors with RGB information to fill in missing depth information for clear objects~\cite{sajjan2020clear} using photorealistic simulation and real data. In contrast to~\cite{sajjan2020clear}, we exclusively use inexpensive RGB sensors and low quality simulated data for training. Our approach is comparable to \citet{xie2020best}, which separately leverages depth and synthetic RGB inputs to compensate for the limitations of depth sensing alone. We compare to several variants of~\citet{xie2020best} as a baseline approach, and find that using stereo images outperforms using RGB and depth on optically challenging objects.
% and a comparatively simple depth reconstruction auxiliary loss and network module, and we show that this is sufficient to extract information to manipulate optically challenging objects.\mike{I world revise how we compare to these two papers; ClearGrasp assumes high quality photo-realism and real data, which we do not. Xie's paper is something that we baseline against and strictly beat on glass objects. It might make sense to break them apart in comparision}

An alternative to depth-based transfer is domain randomization with RGB images (DR)~\cite{tobin2017domain,tremblay2018training,sadeghi2016cad2rl,alghonaim2020benchmarking}. Typically, DR randomizes over the lighting and textures in the environment by utilizing a low-quality, non-photorealistic renderer and has produced successful results in car detection~\cite{tremblay2018training}, 6-DOF pose estimation of known objects~\cite{tremblay2018deep}, and camera calibration~\cite{lee2020camera}.
% One rationalization for the efficacy of DR is that it reduces the feature space by randomizing over texture and lighting. 
% \mike{So this is where our last paper got into trouble. If you read the Cad2RL paper, they argue DR is trying to enforce geometric features. So we should cite them as the main reasoning of how DR works. However, they don't examine how to use stereo to enable robust extraction of geometric feature, which is our insight}
By injecting large randomization into texture and lighting, the network is forced to use the geometry of the scene to solve the task~\cite{sadeghi2016cad2rl}. A limitation of this approach is that geometric reasoning on a monocular image requires leveraging global shape priors, lighting effects, and scene understanding~\cite{howard2012perceiving}. In this work, we propose a transfer technique that uses learned stereo matching with domain randomization to extract geometric features using only local non-semantic information and demonstrate robustness on the KITTI benchmark. 

\subsection{Learned Stereo Matching}
The goal of stereo matching is to try to explicitly compute the pixel displacement offset between objects from the left and right images in a stereo pair~\cite{forsyth2012computer}. By computing a per-pixel displacement offset, or disparity, a high-resolution point cloud can be generated of the world. Given that it is challenging to estimate similarity features of image patches, recent work has explored using deep learning techniques to advance stereo techniques~\cite{luo2016efficient,ilg2017flownet,mayer2018makes,zbontar2015computing,mayer2016large,kendall2017end}.
% The learned stereo community has shown that anaglyphs can be a useful representation for learning feature matching as well as direct regression of coarse depth~\cite{}.
In contrast to the learned stereo community that targets achieving perfect depth-sensing when training with real data~\cite{mayer2016large,kendall2017end}, we are interested in enabling transfer from simulation for ``high-level" vision tasks. We specifically propose a network for manipulation, SimNet, that regresses coarse depth with an efficient stereo matching network architecture based on~\cite{mayer2016large,kendall2017end} to learn features that are robust enough to predict oriented bounding boxes, segmentation masks, and keypoints on real objects for manipulation.

%% file: 3-cost-volume.tex
% \section{\tom{I would call this section panoptic network, and have a subsection for depth reconstruction from SCVNs.  We need to explain the overall approach before we get into stereo, IMO.  This would include all the heads that we have and how the overall loss work.  Citations are fine, but we should re-iterate this at least at a high level.} Depth Reconstruction with \costvolumes{} (\costabbr{}s)}
\section{SimNet: Enabling Predictions for Manipulation  From Synthetic Stereo}

\begin{figure}[t!]
\centering
\includegraphics[width=0.83\linewidth]{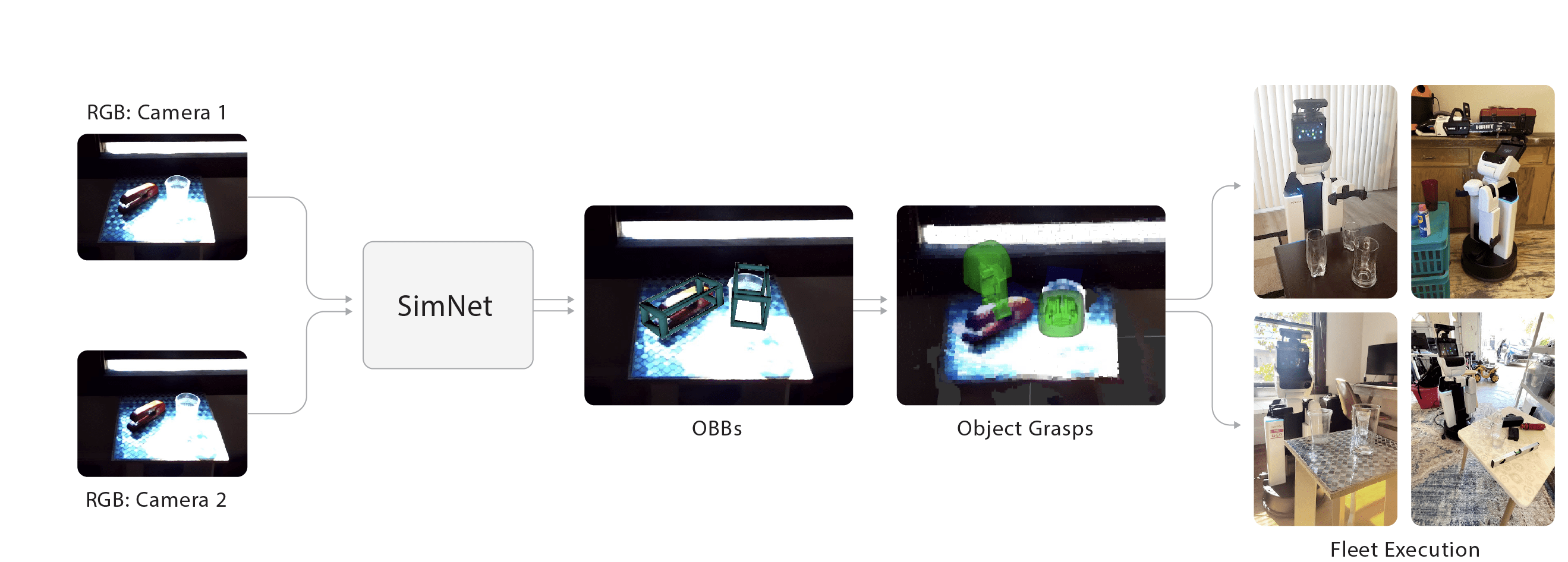}
\caption{\noindent \textbf{Manipulation overview}:  A stereo pair is fed as input to SimNet, which produces OBBs. Grasp positions are produced from OBBs and are used by a classical planner to grasp the object~\cite{balasubramanian2012physical}. This grasping technique has been deployed on a fleet of home robots to perform manipulation in optically challenging scenarios. 
}
\vspace{-4ex}
\label{fig:manipulation}
\end{figure}

\label{sec:stereo_reconstruction}
In this section, we will discuss our network architecture that leverages approximate stereo matching techniques from~\cite{kendall2017end,mayer2016large} and domain randomization~\cite{sadeghi2016cad2rl,tobin2017domain} to predict segmentation masks, OBBs, and keypoints on unseen objects for robot manipulation. Our key insight is that by training approximate stereo matching algorithms on pure synthetic data, we can learn robust ``low-level" features like disparity than enable sim2real transfer of ``high-level" vision tasks. We first discuss how we learn robust low-level features and then feed that into high-level perception algorithms. Finally, we discuss how our low-cost synthetic data is generated. For training details see supplemental Sec.~\ref{app:training_details}; the overall network architecture is in Fig.~\ref{fig:simnet_diagram}.

% \mike{I think we need to be more careful with this section. We aren't inventing a cost volume, since that is a known concept (see https://arxiv.org/pdf/1512.02134.pdf). Our paper is really saying that this technique enables robust sim2real transfer of downstream tasks when trained in combination with the panoptic network. So, it reality this would consider more background than novel research. }

\subsection{\costvolumes{} (SCVN) For Robust Low-Level Features}
% \brijen{from meeting, describe what the feature extractors are}
% Let $\phi_{\rm left}$ and $\phi_{\rm right}$ denote neural networks that featurize the left and right images respectively.
% Each input image is fed into these networks to obtain features $\phi_{\rm left}(I_{\rm left})$ and $\phi_{\rm right}(I_{\rm right})$. Both features have dimension $C \times H \times W$.
We will now describe the stereo matching neural network module used in SimNet's backbone, which enables transfer from simulation. In this section, we use $\odot$ to denote Hadamard products, and use $I_{[i, j:k,:]}$ to select all elements with index $i$ in the first dimension of tensor $I$, index in $\{j,\ldots k -1\}$ in the second dimension of $I$, and any index in the third dimension onwards.
Let $I_{\rm l}$ and $I_{\rm r}$ denote the left and right RGB images from the stereo pair. Each image has dimension $3\times H_0\times W_0$.
The left and right images are fed into neural networks $\Phi_{\rm l}$ and $\Phi_{\rm r}$
that featurize each image respectively and output feature volumes $\phi_{\rm l}$ and $\phi_{\rm r}$. Both $\phi_{\rm l}$ and $\phi_{\rm r}$ have dimension $C_\phi \times H_\phi\times W_\phi$, where $C_\phi$ is the number of channels in each feature volume, and $H_\phi$ and $W_\phi$ are their height and width, respectively. We used a lightweight Dilated ResNet-FPN~\cite{yu2017dilated} as our feature extractor, to enable large receptive fields with a minimal amount of convolutional layers. 

The extracted features $\phi_{\rm l}$ and $\phi_{\rm r}$ are fed into a stereo cost volume network $f_{\rm cost}$ that consists of an approximate stereo matching module that searches horizontally in the feature volumes for correspondences within an allowed disparity range. In classical stereo vision literature, correspondences across left and right images can be found by searching along a horizontal line across the images for a match, and the disparity is the difference in the $x$ coordinates in the match, which is high for closer points in 3D space and low for farther points. The architecture for $f_{\rm cost}$ is heavily inspired by the techniques in~\cite{mayer2016large} and approximately performs this search.
The first phase of the network $f^{(0)}_{\rm cost}$ computes pixelwise dot products between horizontally shifted versions of the feature volumes. The output of this phase has dimension $C_{\rm c} \times H_{\phi} \times W_{\phi}$. $2 * (C_{\rm c}-1)$ represents the maximum disparity considered by the network, and the minimum disparity considered is $0$. The $i$-th $H_{\rm c} \times W_{\rm c}$ slice of the output is computed as:

\begin{align*}
\vspace{-3ex}
    f^{(0)}_{\rm cost}(\phi_{\rm l}, \phi_{\rm r})_{[i,:,i:]} &= \sum_{j = 0}^{C-1}\left(\phi_{\rm l,[i,:,i:]}\odot \phi_{\rm r,[i,:,:W-i]}\right)_{[j]};\hspace{0.2in}
    f^{(0)}_{\rm cost}(\phi_{\rm l}, \phi_{\rm r})_{[i,:,:i]} = 0
\vspace{-3ex}
\end{align*}

The first case takes the rightmost $H_{\rm c} - i$ columns of the left feature volume $\phi_{\rm l}$ and computes a pixelwise dot product with the leftmost $H_{\rm c} - i$ columns of $\phi_{\rm r}$. This operation horizontally searches for matches across the two feature volumes at a disparity of $2i$. The next phase of the network $f^{(1)}_{\rm cost}$ feeds the resulting volume into a sequence of ResNet blocks, which outputs a volume of dimension $C_{\rm c} \times H_{\phi} \times W_{\phi}$ before performing a soft argmin along the first axis of the volume. The soft argmin operation approximately finds the disparity for each pixel by locating its best match. The final volume is an estimate of a low-resolution disparity image $\hat{I}_{\rm d, low}$ with shape $H_\phi \times W_\phi$. We denote $f_{\rm cost} = f^{(1)}_{\rm cost} \circ f^{(0)}_{\rm cost}$.

\noindent \textbf{Disparity Auxiliary Loss}\label{subsec:depth_reconstruction_loss_scvn}
% In Section~\ref{sec:manipulation}, we describe how to use \costabbr{}s to create and train networks that make multi-headed predictions for manipulation policies.
In addition to the losses for the high-level perception heads, we additionally train the weights of $\Phi_l$, $\Phi_r$, and $f_{\rm cost}$ by minimizing an auxiliary depth reconstruction loss function. In particular, the loss function takes in a target disparity image $I_{\rm targ, d}$ of dimension $H_0\times W_0$, downsamples it by a factor of $H_0/H_\phi$ and then computes the Huber loss~\cite{huber1992robust} $\ell_{\rm d, small}$ of it with the low-resolution depth prediction $f_{\rm cost}(\phi_l, \phi_r)$. That is, the network weights are trained to minimize $\ell_{\rm d, small}(f_{\rm cost}(\phi_l, \phi_r), \texttt{downsample}(I_{\rm targ, d}, H_0/H_\phi))$.

\subsection{Extracting High-Level Predictions for Manipulation Tasks}
Given a \costabbr{} to extract geometric features from stereo images, we need to learn on top of it high-level predictions relevant to manipulation.
To design a backbone for robust simulation-trained manipulation, we feed the output of a SVCN, $\hat{I}_{\rm d, low}$ , into a Resnet18-FPN~\cite{kirillov2019panoptic} feature backbone $f_{\rm backbone}$. Additionally, we leverage early stage features from the left RGB image, $I_{\rm l}$, to enable high resolution texture information to be considered at inference time, similar to ~\cite{xie2020best}. The features are extracted from the ResNet stem and  concatenate with the output of the \costabbr{} and fed into the backbone. The output of the backbone is fed into each of the prediction heads. 
% \tom{We should probably have a figure that is a blow up of the images for each of these (e.g., as in Figure 1).}
% The feature extractor $f_{\rm backbone}$ is then fed into several possible output heads to output representations for manipulation policies.
In this section, we describe each of the prediction heads of SimNet (Fig.~\ref{fig:simnet_diagram}) and the losses used to train it.  Each prediction head high uses the up-scaling branch defined in~\cite{kirillov2019panoptic}, which aggregates different resolutions across the feature extractor.  

\paragraph{Room Level Segmentation:}
Given a mobile robot that is trying to manipulate objects on a table, it is useful to be able to know where the table is in the room and the objects on the table. We can express this level of scene understanding as segmentation problem consisting of predicting three categories: surfaces, objects and background.  Cross-entropy loss $l_{\rm seg}$ is used for training, as in prior work~\cite{kirillov2019panoptic}. 

\paragraph{Oriented Bounding Boxes:}
Detection of an OBB requires determining individual object instances as well as estimating translation, $t \in \mathbb{R}^3$, scale $S \in \mathbb{R}^{3 \times 3}$, and rotation, $R \in \mathbb{R}^{3 \times 3}$, of the encompassing OBB. We can recover these parameters by using four different output heads. First to recover object instances, we regress a $W_0 \times H_0$ image,which is the resolution of the input left image, where for each object in the image a Gaussian heatmap is predicted. Instances can then be derived using peak detection as in ~\cite{hou2020mobilepose, duan2019centernet}. We use an $L_1$ loss on this output head and denote the loss as $l_{\rm inst}$.

Given instances of object, we can now regress the remaining 9-DOF pose parameters. To recover scale and translation, we first regress a $W_0/8 \times H_0/8 \times 16$ output head where each element contains pixel-wise offset from detected peak to the 8 box vertices projected on to the image. We can then recover the scale and translation of the box up to a scale ambiguity using EPnP similar to ~\cite{hou2020mobilepose}. In contrast to prior work in pose estimation, we aligned the predicted box based on principal axes sized in a fixed reference frame. To recover absolute scale and translation, we additionally regress the distance from the camera $z \in \mathbb{R}$ of the box centroid as a $W_0/8 \times H_0/8$ tensor. The two losses on these tensors are an $L_1$ loss and are denoted $l_{\rm vrtx}$ and $l_{\rm cent}$.

Finally the rotation, $R$ of the OBB can be recovered via directly predicting the covariance matrix, $\Sigma \in \mathbb{R}^{3 \times 3}$ of the ground truth 3D point cloud of the target object, which can be easily generated in simulation. We directly regressed an output tensor of $W_0/8 \times H_0/8 \times 6$, which for each pixel contains both the diagonal and symmetric off diagonal elements of the target covariance matrix. We can then recover rotation based on the \textbf{SVD} of $\Sigma$. We additionally use an $L_1$ loss on this output head and denote the loss $l_{\rm rot}$. Note for the 9-DOF pose loses, we only enforce them where the Gaussian heatmaps are greater than $0.3$, to prevent ambiguity in empty space.

\paragraph{Keypoints:}
Keypoints and learned correspondences are a common representation for robot manipulation, especially in deformable manipulation~\cite{sundaresan2020learning,grannen2020untangling,ganapathi2020learning,manuelli2019kpam,seita2018deep}. SimNet has an output head that predicts keypoints of various classes, which can be fed into planners for manipulation. For instance, one of the keypoint classes we predict in Sec.~\ref{sec:experiments} are t-shirt sleeves, which is used for t-shirt folding. The output head predicts heatmaps for each keypoint class, and is trained to match target heatmaps with Gaussian distributions placed at each ground-truth keypoint location using a pixelwise cross-entropy loss $l_{\rm kp}$. To extract keypoints from the predicted heatmaps, we use non-maximum suppression to perform peak detection as in~\citet{hou2020mobilepose}.
% \brijen{Given that there are many ways to formulate keypoint prediction, I think we should keep the specific details abstract here and specify the exact losses used in the appendix.}

\paragraph{Full Resolution Disparity:}
The final output head is a full resolution disparity image, which can be converted into a 3D point cloud for collision avoidance. Since our \costabbr{}, produces a disparity image at quarter resolution, we can combine the feature extractor backbone and the left stereo image to produce a full resolution depth image. We use the same branch architecture as the previous heads to aggregate information across different scales. During training we use the same loss as the \costabbr{}, but enforced at full resolution. This output is trained using a Huber loss function.

\subsection{Efficient Synthetic Dataset Generation}
% \noindent \textbf{Synthetic Data}
Given the complexity of all the output predictions defined in the previous section, it would be impractical to label a sufficient amount of real-data to generalize across scenes. Thus, we are interested in using synthetic data to provide ground truth annotations on a wide variety of scenarios. To force the network to learn geometric features, we randomize over lighting and textures as recommended in~\citet{mayer2018makes}. In contrast to~\cite{sajjan2020clear,roberts2020hypersim}, we use OpenGL shaders with PyRender~\cite{matlpyrender} instead of physically based rendering~\cite{pharr2016physically} approaches. Low-quality rendering greatly speeds up computation, and allows for data-set generation on the order of an hour. We generate three datasets: cars, graspable objects, and t-shirts (Fig.~\ref{fig:simulator}), and discuss specific dataset details in supplemental Sec.~\ref{app:dataset}. 
We intend to release our datasets as well as real-world validation data as part of this paper. 

% \noindent \textbf{Optimization} For an input training stereo image $(I_{\rm l}, I_{\rm r})$ and corresponding labels, the network is trained by minimizing $\lambda_{\rm seg}\ell_{\rm seg} + \lambda_{\rm kp}\ell_{\rm kp} + \lambda_{\rm d}\ell_{\rm d} + \lambda_{\rm d}\ell_{\rm d, small} + \lambda_{\rm cov}\ell_{\rm cov}+\lambda_{\rm inst}\ell_{\rm inst} + \lambda_{\rm vrtx} + \lambda_{\rm cent}$. We provide the loss weights $\lambda$ in \brijen{ref}.The network is trained end-to-end using the Adam optimizer~\cite{kingma2014adam} and Batch Normalization~\cite{ioffe2015batch} between each layer on the datasets described in Sec. \ref{sec:simnet}. Our network is trained on a single Tesla V100 GPU with a batch size of 18 images. To tune the loss weights, we use HyperBand~\cite{li2017hyperband} across 20 single gpu instances for 48 hours. 

% The network is trained on simulated images rendered using PyRender~\cite{matlpyrender}. To force the network to learn geometric features, we randomize over lighting and textures as in~\citet{mayer2016large}. In contrast, we use OpenGL shaders instead of physically-based rendering, and found this to be sufficient despite its much less realistic appearance (Figure~\ref{fig:simulator}). This additionally speeds up rendering by several orders of magnitude. We release the following datasets consisting of stereo images from the simulator and small sets of real data with annotations. The simulated images have annotations for segmentation masks, depth values, bounding boxes, and keypoints.
% \section{SimNet: A Lightweight Domain Randomized Simulator}
\label{sec:simnet}

%% file: 5-experiments.tex
\begin{figure}[t!]
    \centering
    \includegraphics[width=5.5in]{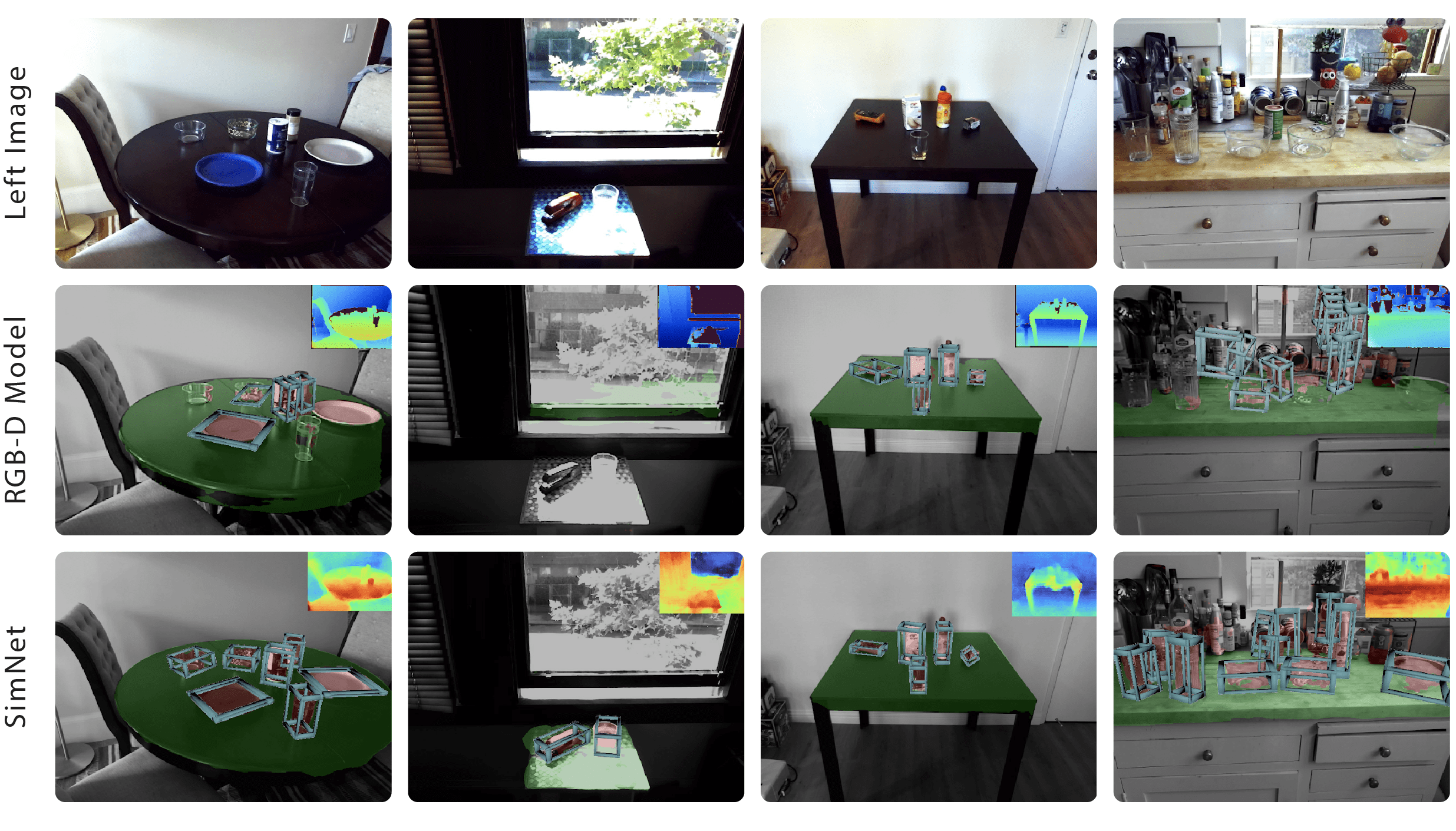}
\caption{\textbf{Graspable Object Predictions:} \net{} is evaluated on an oriented bounding box regression task with objects of varying sizes and shapes placed on flat surfaces. Top: Is the left RGB image from different homes. Middle: Is our model trained with RGB-D sim2real transfer similar to ~\cite{xie2020best}. The top right corner is the output of the Asus Xtion depth sensor. Bottom: Is SimNet. In the top right, we show the low-res disparity estimate predicted by the learned stereo net. SimNet consistently enables better sim2real transfer of the predictions for optically challenging scenarios.}
\label{fig:grasppredictions}
\vspace{-3ex}
\end{figure}

\section{Experiments}
SimNet is evaluated on a set of real-world computer vision and robotics tasks to see if training on synthetic stereo data will transfer robustly to diverse, real images in unstructured environments and enable manipulation of optically challenging objects. SimNet is compared against baseline approaches using monocular, naive stereo concatenation and active RGB-D sensing (supplemental Sec.~\ref{app:baselines}).

\subsection{Manipulation Experiments}
% \tom{IMO, we should lead with this since grasping unknown objects is our main contribution and grasping optically challenging unknown objects is where we shine.  On the t-shirt folding side, why do we think we are better than other t-shirt folding approaches?  Sim2real? Panoptic-predictions?  A combination?}
\label{sec:experiments} 
% We evaluate the ability of the simulation-trained stereo reconstruction network to transfer to real home data generated during manipulation of optically challenging objects.
The learned perception models and manipulation policies are evaluated on two optically challenging tasks: (1) grasping objects on tabletops and (2) t-shirt folding.
\subsubsection{Experimental Setup}
All physical experiments are conducted on tabletops found across four different, real homes. Manipulation is performed with the Toyota HSR robot, which has a four DOF arm, mobile base, and pair of parallel jaw grippers~\cite{yamamoto2019development}. It has a mounted stereo pair from a Zed 2 camera and a Asus Xtion Pro RGB-D sensor. Each home has different background objects, furniture, graspable objects, and lighting conditions, which evaluates each network's ability to robustly generalize to diverse scenarios.
% We conduct experiments in two different lighting conditions: direct sunlight and medium lighting. \brijen{better word for this}
\begin{figure}[t!]
    \centering
    \includegraphics[width=5.5in]{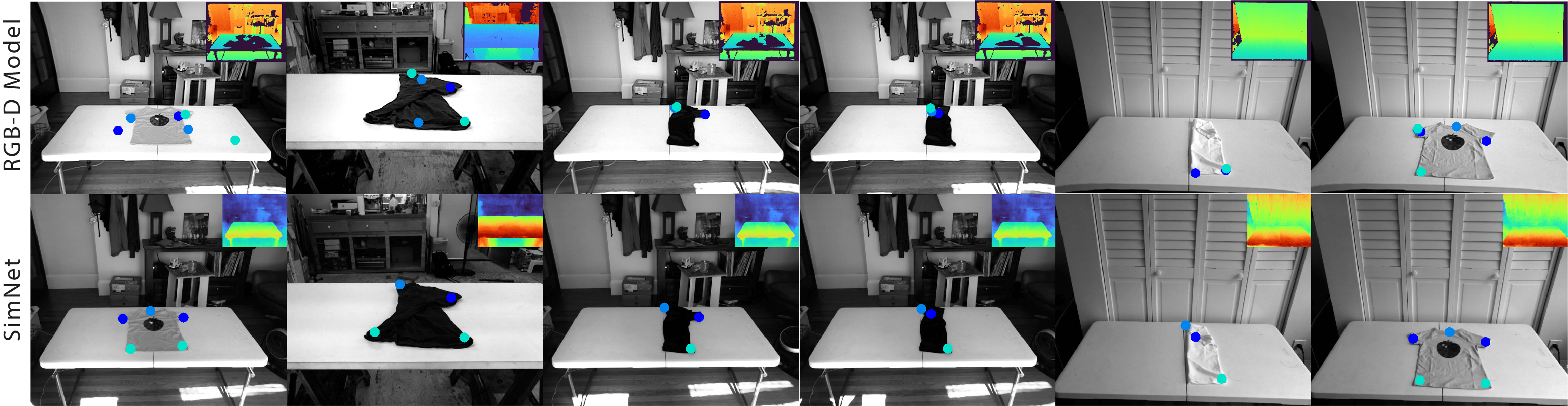}
\caption{\textbf{T-shirt Keypoint Predictions:} \net{} is evaluated on keypoint regression for shirts in various stages of folding. Three classes of keypoints are predicted: sleeves, neck, and bottom corners. RGB-D performs poorly due to strong natural lighting and minimal depth variation. \net{} accurately predicts keypoints on the shirts despite these challenges.}
\label{fig:shirtpredictions}

\end{figure}

\subsubsection{Ablation of Baselines on Validation Data} \label{subsubsec:panoptic_ablation}
We compare to the following baselines: \textbf{mono}, which uses the left stereo image, \textbf{depth}, which only uses depth inputs, and \textbf{RGB-D}, which uses both RGB and depth inputs. To make the RGB-D baseline competitive, we swept over four different algorithms to fuse color and depth information on a held out set of $500$ non-optically challenging scenes (i.e. matte objects with minimal natural light). We measured 3D mAP@0.25, a common metric in pose prediction~\cite{wang2019normalized}, of annotated bounding boxes of each object on the tabletop of interest. We report the best results of each method in Table \ref{table:panoptic_perception_results}. 
% \begin{enumerate}
%     \item \textbf{Monocular (Mono):} this network only views the left image from the Zed 2.
%     \item \textbf{Depth:} this network only views the depth image from the PrimeSense.
%     \item \textbf{RGB-D:} this network uses both RGB and depth from the PrimeSense as input.
% \end{enumerate}`
Additional sensing ablations can be found in Sec.~\ref{app:ablation}.

\begin{table}[!htbp]
\centering
% \vspace{-0.2cm}
% \resizebox{\columnwidth}{!}{
% density, knot, rand success, rand actions, hulk success, hulk actions, hulk failures
 \begin{tabular}{| c | c | c | c | c |}
\hline
Method & Mono & Depth & RGB-D & SimNet \\ 
\hline
3D mAP & 0.164 & 0.831 & 0.855 & \textbf{0.921}\\
% Depth & \\
% RGB-D & 0.855\\
% \costabbr{} & \textbf{0.921}\\
\hline
\end{tabular}
% }

\caption{\textbf{SimNet Perception Results:} We present an ablation of different sensing modalities on the 3D OBB prediction task. We evaluate 3D mAP on a dataset of real, human-annotated images of optically \textbf{easy objects.} We find that monocular sensing performs poorly on this task, likely due to the task being 3D in nature, while depth and RGB-D perform much better. SimNet outperforms these methods on the real images. Baseline implementation details can be found in Sec.~\ref{app:baselines}. We present full ablation results in supplemental Sec.~\ref{app:ablation}.}
\label{table:panoptic_perception_results}

\end{table}

% \brijen{Talk about robot, camera, number of homes, setup within the homes.}
\subsection{Unknown Object Grasping}
In this experiment, the robot's task is to grasp objects on a tabletop comprised of two classes of household objects in each home: \textbf{optically easy}, which consists of opaque, non-reflective objects, and \textbf{optically hard}, which contains optically challenging, transparent objects (Fig.~\ref{fig:grasping_setup}). For each trial, we select an object uniformly at random from the dataset and randomly place it on the tabletop with other distractor objects. The task is to grasp the foremost object in the scene, by using a heuristic grasp planner that takes OBBs as input. Specifically, given an OBB, the robot aligns the gripper with the largest principal axis. In the event, of similar sized principal axes like a ball, the robot favors grasping the object on the side closest to the robot.  A grasp is successful if the robot is able to raise the object off the table and remove it from the scene. 

For each of the four homes we test five easy objects and five hard objects and compare SimNet against the best RGB-D baseline found in Sec. \ref{subsubsec:panoptic_ablation}.
We report quantitative results in Table \ref{table:grasping_results} and qualitative results of the predictions in Fig. \ref{fig:grasppredictions}. SimNet outperforms RGB-D, 92.5\% vs. 62.5\% in grasp success. 
% The perception network is trained on domain-randomized, simulated stereo images as in Figure \brijen{todo}. The depth reconstruction features are fed into output heads for semantic segmentation mask prediction and oriented bounding box (OBB) detection. The robot generates grasps from OBBs by
% \brijen{describe manipulation policy}

% \subsubsection{Evaluation Metrics:}

\begin{table}[!htbp]
\centering
% \vspace{-0.2cm}
% \resizebox{\columnwidth}{!}{
% density, knot, rand success, rand actions, hulk success, hulk actions, hulk failures
 \begin{tabular}{| c | c |c|c|c|c|}
\hline
Method (Object Class) & Home 1 & Home 2 & Home 3 & Home 4 & Overall\\ 
\hline
RGB-D (O. Easy) & 4/5 & \textbf{5/5} & 4/5 & \textbf{5/5} & \textbf{18/20}\\
\net{} (O. Easy) & \textbf{5/5} & 4/5 & \textbf{5/5} & 4/5& \textbf{18/20}\\
RGB-D (O. Hard) & 0/5 & 1/5 & 1/5 & \textbf{5/5} & 7/20\\
\net{} (O. Hard) & \textbf{5/5} & \textbf{5/5} & \textbf{5/5} & 4/5 & \textbf{19/20}\\
\hline

\end{tabular}
% }

\caption{\textbf{SimNet Grasping Results:} Grasp success scores between the best RGB-D method and SimNet across homes. On  optically easy objects (i.e. matte and non-reflective) RGBD-D and SimNet perform similar on average. However, when presented with challenging objects such as glassware, SimNet outperforms RGB-D.}
\label{table:grasping_results}

\end{table}

% \brijen{task name should also imply that grasping + rearrangement are necessary}

\subsection{Grasp Point Prediction in T-shirt Folding}
The robot is also evaluated on a t-shirt folding task, where it must execute a sequence of four folds on unseen, real t-shirts. This task is challenging to perform using depth sensing, because the depth resolution of most commercial depth sensors cannot capture the subtle variations in depth due to the thickness of a t-shirt. Keypoints are a popular representation for manipulating deformable objects~\cite{grannen2020untangling,seita2018deep}. We parameterize a robot shirt folding policy using keypoint predictions for the shirt's neck, sleeves, and bottom corners (Sec.~\ref{app:t-shirt-manip}). To compute quantitative results on sim2real transfer, we collect a validation dataset of 32 real images from 12 t-shirts in 3 homes of stages of t-shirt folding (Fig.~\ref{fig:shirtpredictions}) and report keypoint prediction mAP (Table~\ref{table:shirt_results}). We additionally present videos of folding using the robot on the project website for qualitative evaluation. We find that \net{} significantly outperforms RGB-D and depth, and slightly outperforms mono. A large part of our dataset was collected in a room with direct natural light, which increase noise in the active depth sensor, due to large amount of infrared light. 

% To predict the location of the grasp and place actions, we use the stereo depth reconstruction network's features as input to a keypoint regression head. For each image, the keypoint regression network predicts four keypoint classes: neck, sleeves, bottom corners, and the point on the bottom edge of the shirt that is leftmost in the scene. These keypoints are used to parameterize the robot policy that executes each fold in sequence.

\begin{table}[!htbp]
\centering
% \vspace{-0.2cm}
% \resizebox{\columnwidth}{!}{
% density, knot, rand success, rand actions, hulk success, hulk actions, hulk failures
 \begin{tabular}{| c | c | c | c | c |}
\hline
Method & Mono & Depth & RGB-D  & SimNet \\ 
\hline
mAP & 0.893  & 0.282 & 0.631 & \textbf{0.917} \\
% Stacked & 0.710\\
% SimNet & \textbf{0.796}\\
\hline
\end{tabular}
% }

\caption{\textbf{T-Shirt Folding Results:} A comparison of keypoint mAP shows that active depth-based models transfer poorly on this task, due to interference from natural lighting and low depth profile of the shirts. The monocular network transfers well to the real shirt images, and \net{} slightly outperforms it. Unlike the models that use active depth sensing, \net{} is robust to lighting variation and the low depth variation of the shirts.}
\label{table:shirt_results}

\end{table}

% \subsubsection{Evaluation Metrics:}

\begin{figure}[t!]
    \centering
    \includegraphics[width=5.5in]{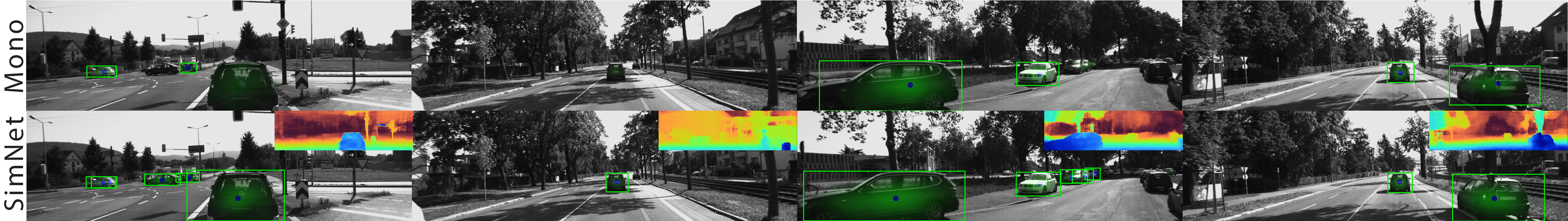}

\caption{\textbf{KITTI bounding box predictions:} We evaluate SimNet on the 2D bounding box detection task in the KITTI benchmark. Top: Is the prediction of monocular-DR. Bottom: The predictions of SimNet with the estimated disparity from the SVCN in the top right corner. By performing approximate stereo matching SimNet achieves high-prediction accuracy via improved transfer from simulation. 
% \mike{I think this is perfect, can you make sure to use the svcnn depth and have the images aligned correctly (i.e. top-left svcnn should be the same as top-left mono}
% \brijen{Honestly, we might want to move this fig to the appendix}
}
\label{fig:kittipredictions}

\end{figure}

\subsection{KITTI 2D Car Detection}
% \tom{We need an excuse to have this here.  For example, we would ideally say that our approach is pretty close to state of the art on kitti, but IIRC that wasn't the case.} \brijen{from meeting: better motivate this benchmark}
\label{sec:kitti}
In this experiment, we evaluate whether \net{} is effective sim-to-real for other  vision tasks outside of home robotics such as 2D car  detection on the KITTI  benchmark~\cite{Geiger2012CVPR,Geiger2013IJRR}. We compare the stereo cost volume network trained on simulated images to networks using monocular inputs (\textbf{Mono}), stacking left and right image inputs (\textbf{Stacked}), and using real data with a standard validation/training split used in~\cite{duan2019centernet} (\textbf{SimNet-real}). To predict 2D bounding boxes, we modify SimNet to use the single-shot box detection output head in CenterNet~\cite{duan2019centernet}. In the table below we report mAP@0.5 for the aggregate of moderate and easy classes car classes. 

Interestingly, SimNet can significantly outperform monocular images and naive stereo concatenation, which suggests explicit stereo matching leads to more robust transfer. Furthermore, the gap between real and sim data is only 3.5\%, which suggests relatively robust transfer. However, qualitatively SimNet fails to detect cars at distance, since the ability to reason about geometry decays with distance from the camera. For visualizations of the predictions see supplemental Fig.~\ref{fig:kittipredictions}. Thus, SimNet is best used for applications with limited range from the camera, such as robot manipulation. 

\begin{table}[!htbp]
\centering
% \vspace{-0.2cm}
% \resizebox{\columnwidth}{!}{
% density, knot, rand success, rand actions, hulk success, hulk actions, hulk failures
 \begin{tabular}{| c | c | c | c || c |}
\hline
Method & Mono & Stacked & SimNet & SimNet-real \\ 
\hline
mAP & 0.565 & 0.710 & \textbf{0.826} & 0.861 \\
% Stacked & 0.710\\
% SimNet & \textbf{0.796}\\
\hline
\end{tabular}
% }

\caption{\textbf{KITTI  Results:} Results of different sim-to-real techniques for 2D car detection on the KITTI Benchmark. We find that the simply stacking the left and right images in the stacked input network significantly outperforms the monocular network. SimNet, however, transfers much better than naive stereo concatenation and is close to performance using real data.}
\label{table:kitti_results}

\end{table}

%% file: 6-conclusion.tex
\section{Conclusion}
\label{sec:conclusion}
We present \net{}, am efficient, multi-headed prediction network that leverages approximate stereo matching to transfer from simulation to reality. The network is trained on entirely simulated data and robustly transfers to real images of unknown optically-challenging objects such as glassware, even in direct sunlight. We show that these predictions are sufficient for robot manipulation such as t-shirt folding and grasping. In future work, we plan to use \net{} to automate household chores.

\section{Acknowledgements}
We like to thank Jeremy Ma for designing our stereo data collection rig used to collect indoor scenes.  We also thank Krishna Shankar and Max Bajracharya for their invaluable insights into both classical and learned stereo techniques. Finally, we like to thank the Machine Learning Research team at TRI for their  feedback and guidance.  

%% file: 7-appendix.tex
\newpage
\appendix

\section{Stereo Data Collection Rig}
We now describe the two physical rigs used to collected stereo images for validation. In each rig, there is also an RGB-D sensor that is also used to collect data to evaluate the depth-based baselines.
\subsection{Setup 1: Basler Stereo Pair}\label{app:basler}
To collect data in the homes, we created a a stereo pair with two Basler 60uc cameras. Our camera's native resolution is $2560 \times 2048$ pixels with a wide field of view fish-eye lens. However, we perform inference on a 4x down-sampled version.  To obtain a stereo pair, all images are rectified to a pinhole camera model. The baseline of our stereo camera is $10cm$, which was selected to match the average  distance between human eyes.  

\begin{figure}[t!]
    \centering
    \includegraphics[width=5.5in]{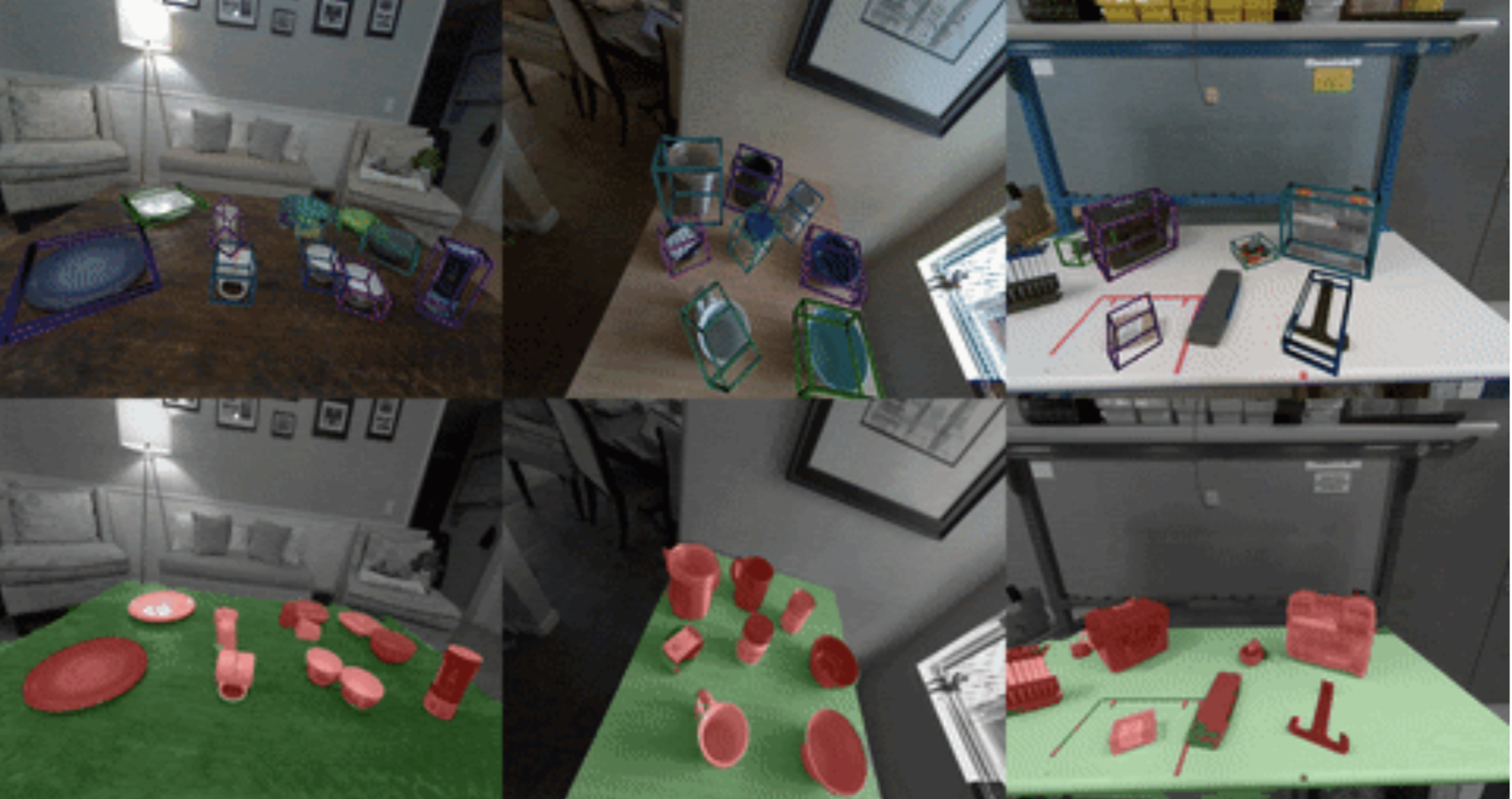}

\caption{\textbf{\net{} Predictions on Indoor Scenes:} We present visualizations of \net{}'s predictions on the dataset used in the sensing ablation studies. This dataset consists of optically easy scenarios , but can have many objects in the each scene. We find that the OBB and scene level segmentation predictions from SimNet on this dataset are high quality compared to baselines (Section~\ref{app:ablation}).}
\label{fig:grasping_basler}
\end{figure}

To obtain depth data for our baseline technique, we mount a Microsoft Kinect Azure in wide field of view mode. The Kinect is mounted to a fixed steel base on top of stereo camera pair. We calibrate the Kinect to the left stereo camera using standard checkerboard calibration~\cite{forsyth2012computer}. Camera calibration allows us to project the Kinect's point cloud into the left camera.  We down sample the depth data  using a $2\times 2$ median filter to match the resolution inference is performed at. Example images  from our data collection rig can be seen on Fig. \ref{fig:grasping_basler}.

\subsection{Setup 2: HSR Stereo Pair}\label{app:zed_Asus Xtion}
On the physical experiments run on the Toyota HSR and for the t-shirt experiments, we the stereo pair from a mounted Zed2 camera, which has resolution $1920\times 1024$ pixels. We down-sample images by a factor of 2, so the images used for training and prediction have resolution $960\times 512$ pixels. The Zed2 camera is a time synchronized stereo pair camera with a baseline of $12cm$. The depth and RGB-D experiments on the HSR are run using a mounted ASUS Xtion, which is placed $5cm$ below the Zed2. 

% \section{Experimental Visualizations}\label{app:exp_visuals}
% We present additional visualizations of predictions for each task in this section.
% \subsection{Object Grasping Predictions}
% We present the object grasping setup and object classes for each of the 4 homes in Figure~\ref{fig:grasping_setup}. We present predictions of the RGB-D network and \net{} in Figures~\ref{fig:grasping_vis_rgbd} and~\ref{fig:grasping_vis_stereo}. We observe that the RGB-D sensor consistently fails to detect the transparent objects at all in the depth images. However, the learned disparity prediction outputs coarse predictions even for transparent objects, despite no transparent objects being in the training set. Even though the RGB-D network still has color inputs, it fails to predict OBBs or segmentation masks for the hard objects. We hypothesize that depth sensing, either via active sensing or implicitly via stereo vision, is necessary to perform this task robustly. This is further corroborated by ablations in Section~\ref{app:ablation} that suggest that monocular RGB input alone is insufficient for this task.

% We additionally present predictions of the network on the easy validation dataset used in the ablation studies in Figure~\ref{fig:grasping_basler}. This images in this dataset contain many more household objects in each scene than the images used in the grasping experiments. However, all of the objects are fairly easy to perceive via active sensing as well. We find that \net{} performs very well on this dataset in our ablation studies (Section~\ref{app:ablation}).

\begin{figure}[t!]
    \centering
    \includegraphics[width=5.5in]{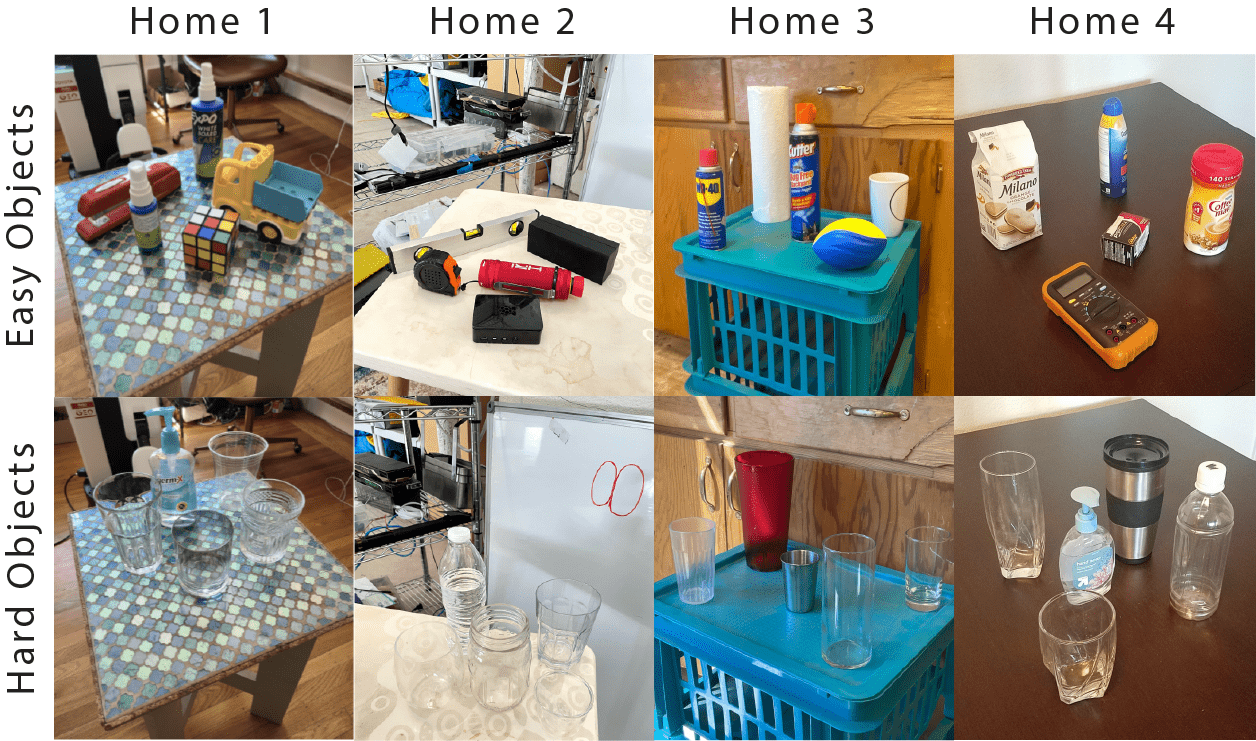}

\caption{\textbf{Grasping Objects and Experimental Setup:} We split objects in each house into two categories: easy and hard. Hard objects are typically transparent, reflective, or translucent, making them hard to perceive with active depth sensors. Each house has 5 objects in each class. We also present the experimental setup in each house for these experiments. Each house has different lighting conditions, tables, and background objects. During grasping experiments, two objects or more are placed on the table, and the robot grasps and extracts the object located in front.}
\label{fig:grasping_setup}
\end{figure}

% \begin{figure}[t!]
%     \centering
%     \includegraphics[width=5.5in]{figures/mike_grasping_images/rgbd_U8sbwbUh4tZSC22DQTF5J3.png}
%     \vspace{0.2in}
%     \includegraphics[width=5.5in]{figures/mike_grasping_images/rgbd_ccbueWB8cXRjUNbxtvpqDR.png}
% \caption{\textbf{RGB-D Grasping Network Predictions:} We observe that the RGB-D network using the Asus Xtion camera is able to predict OBBs, segmentation masks, and depth for the opaque bottles (top). However, it fails when the objects are transparent, and even the segmentation masks for the table's surface are not detected (bottom). We visualize depth images for the RGB-D baseline and predicted disparity images for the \costabbr{}, which explains the variation in color scheme.
% }
% \label{fig:grasping_vis_rgbd}
% \end{figure}

% \begin{figure}[t!]
%     \centering
%     \includegraphics[width=5.5in]{figures/mike_grasping_images/stereo_H52HdtHPWY5RQS9TTdiCnR.png}
%     \vspace{0.2in}
%     \includegraphics[width=5.5in]{figures/mike_grasping_images/stereo_f3WYmaSohpXsDWKcRf8YZf.png}
% \caption{\textbf{\net{} Grasping Network Predictions:} We observe that \net{} using the Zed 2 camera is able to predict OBBs, segmentation masks, and disparity for the opaque bottles (top) and transparent objects (bottom). The clear glass is detected in the learned stereo disparity image, whereas it is not in the RGB-D network (Figure~\ref{fig:grasping_vis_rgbd}).
% }
% \label{fig:grasping_vis_stereo}
% \end{figure}

% \subsection{T-shirt Folding}

\section{Implementation Details}

% \subsection{Network Architecture}

% \subsection{Task Loss Functions}
% In this section, we describe how the individual prediction heads are trained on target labels.

% \subsubsection{Full Resolution Disparity Prediction}
% To train the \net{} full-resolution disparity image prediction head, we use a Huber loss function with respect to a target full resolution disparity image. This is similar to the procedure used to train the low-resolution disparity prediction from the \costabbr{} stereo matching module.

% \subsubsection{Scene Level Segmentation Prediction}
% To train the scene level segmentation head, we create target mask images for each class. We train the output channel of the segmentation prediction head using a pixelwise binary cross-entropy loss function on its corresponding target mask image.

% \subsection{Oriented Bounding Box Prediction}

% \subsubsection{Keypoint Regression}
% In the t-shirt folding task, the robot policy requires keypoint predictions of semantically-relevant points on the t-shirt. In a training left image, for each keypoint class (e.g. neck), we construct a target heatmap by placing Gaussian heatmaps centered at each instance of the keypoint in the image. We then use a pixelwise binary cross-entropy loss function $l_{\rm kp}$ to train the keypoint prediction head to match the targets for each keypoint class.

\subsection{Baseline Implementation Details}\label{app:baselines}
In this section, we will describe implementation details for the baselines used.

\begin{figure}[t!]
    \centering
    \includegraphics[width=5.5in]{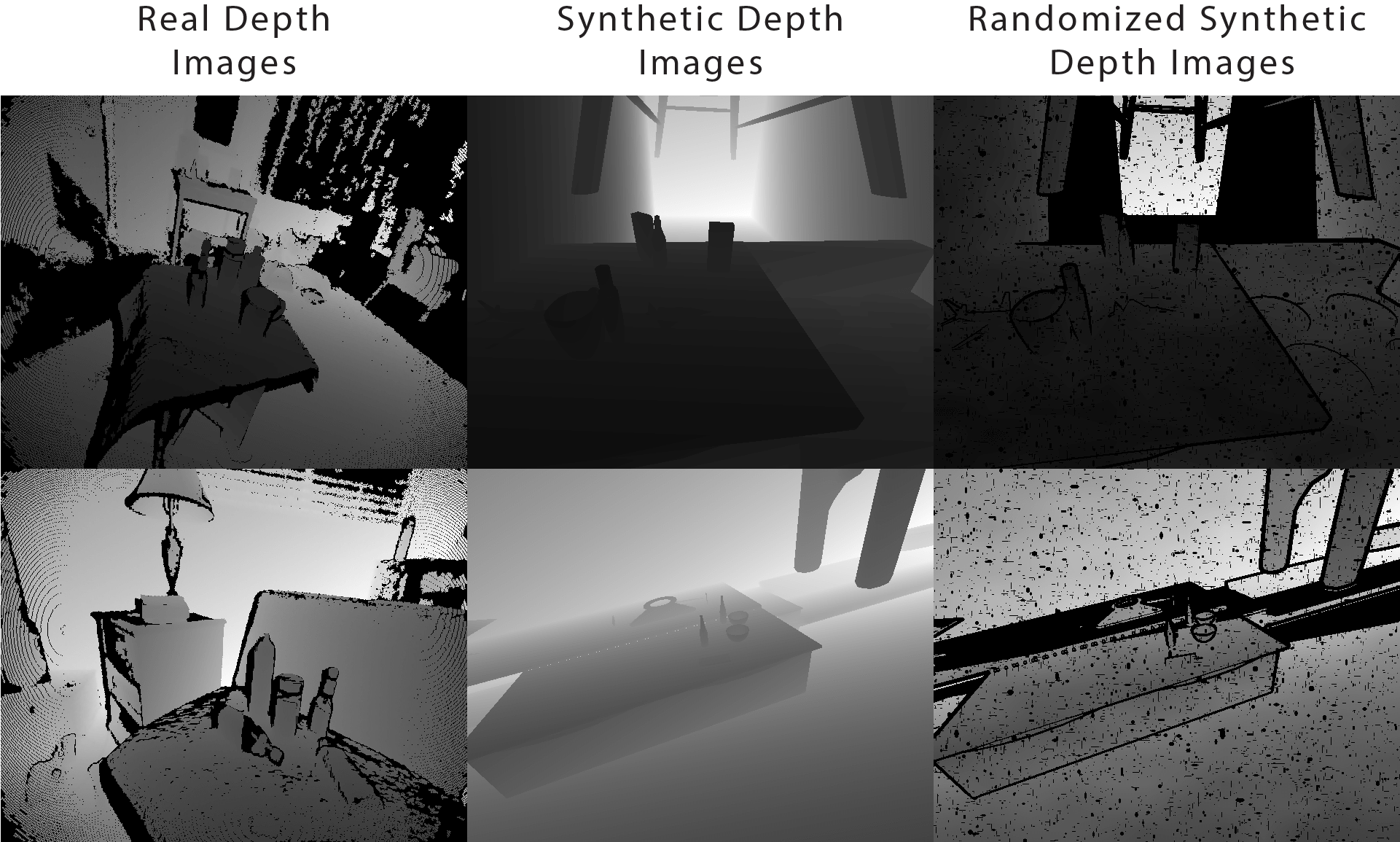}

\caption{\textbf{Depth Image Noise Injection:} To simulate the artifacts present in real depth images (right column), we take synthetically generated depth images (middle column), and add process noise and random dropouts, particularly around edges, similar to~\citet{mahler2019learning}.}
\label{fig:depth_noise}
\end{figure}

\begin{enumerate}
    \item \textbf{Monocular (Mono):} This network trains only on the left image from the stereo pair. The left image is featurized as in the fully stereo network to form $\phi_{\rm l}$ and then fed into the backbone.
    \item \textbf{Monocular, Auxiliary Loss (Mono-Aux):} This network trains only on the left image from the stereo pair. In contrast to Mono, this network takes one of the channels of $\phi_{\rm l}$ and applies the disparity reconstruction auxiliary loss, described in Section~\ref{subsec:depth_reconstruction_loss_scvn}. This forces the monocular network to reason about geometry in the scene.
    \item \textbf{Depth:} This network only trains on depth images, and is identical to the Mono network, except for the number of input channels. Because depth images from real sensors contain many artifacts, we process each synthetic depth image by adding noise and random ellipse dropouts, as in ~\citet{mahler2019learning}. This process is illustrated in Figure~\ref{fig:depth_noise}.
    \item \textbf{Depth, Auxiliary Loss (Depth-Aux):} This network only trains on depth images and is identical to the Mono-Disp network, except for the number of input channels (1). We use the auxiliary loss from Section~\ref{subsec:depth_reconstruction_loss_scvn} on the last channel of the predicted features to predict a low-resolution depth image (not disparity image), before feeding the feature volume to the ResNet50-FPN backbone. This forces the network to try to clean up artifacts in the real or noisy depth images.
    \item \textbf{RGB-D:} This network separately featurizes the RGB and depth images using seperate ResNet stems. It then concatenates computed feature channels and feeds them into the ResNet50-FPN backbone.
    \item \textbf{RGB-D, Auxiliary Loss (RGB-D-Aux):} This network separately featurizes the RGB and depth images, concatenates the features and feed them into the ResNet50-FPN backbone. However, on one channel of the depth features, we apply the depth reconstruction auxiliary loss function (Section~\ref{subsec:depth_reconstruction_loss_scvn}).
    \item \textbf{RGB-D, Stacked Input (RGB-D-Stack):} This network concatenates the RGB and D channels of the input into a single $H\times W\times 4$ array and feeds it into the network as a single input image. 
    \item \textbf{RGB-D, Sequential Input (RGB-D-Seq):} This baseline is heavily inspired by~\cite{xie2020best}, and first trains a depth network as described above. The segmentation and disparity prediction outputs are then fed as input to a mono network in addition to the left image of the stereo pair. The idea with the network architecture is to use RGB information to simply refine the output of the depth prediction network. The depth network is frozen after is trained, and is not updated when the refinement network is trained.
\end{enumerate}

\subsection{Training Details}\label{app:training_details}
All networks are trained on a Telsa V100 GPU for 400,000 gradient steps with a batch size of 18. Models are trained using the Adam optimizer with learning rate $\alpha = 5e-4$ and moment estimate decay rates $\beta_1 = 0.9$ and $\beta_2 = 0.99$. 
The network is trained end-to-end using Batch Normalization~\cite{ioffe2015batch} between each layer on the datasets described in Sec. \ref{sec:simnet}. We trained the network from scratch with random weights. 

For an input training stereo image $(I_{\rm l}, I_{\rm r})$ and corresponding labels, the network is trained by minimizing $\lambda_{\rm seg}\ell_{\rm seg} + \lambda_{\rm kp}\ell_{\rm kp} + \lambda_{\rm d}\ell_{\rm d} + \lambda_{\rm d}\ell_{\rm d, small} + \lambda_{\rm cov}\ell_{\rm cov}+\lambda_{\rm inst}\ell_{\rm inst} + \lambda_{\rm vrtx}\ell_{\rm vrtx} + \lambda_{\rm cent}\ell_{\rm cent}$. To tune the loss weights, we use HyperBand~\cite{li2017hyperband} across 20 single gpu instances for 48 hours. 
% Given the large number of output head performing different tasks, we performed a random search over different loss weights across \brijen{Mike:insert} training instances on EC2 with single TitanV GPUs per training job.  While this is computationally expensive during training, multi-head predictions enable efficient  prediction of all output heads at inference time.
On a TITANXp GPU, our network can predict room level segmentation, OBBs of unknown objects, key-points and full resolution depth at 20 Hz.

% \brijen{I would move this section to the appendix, thoughts?}

\subsection{Evaluation Criteria}\label{app:evaluation}
In this section we will describe how mAP scores are computed for the 3D OBB and keypoint prediction tasks.

\begin{figure}[t!]
    \centering
    \includegraphics[width=5.5in]{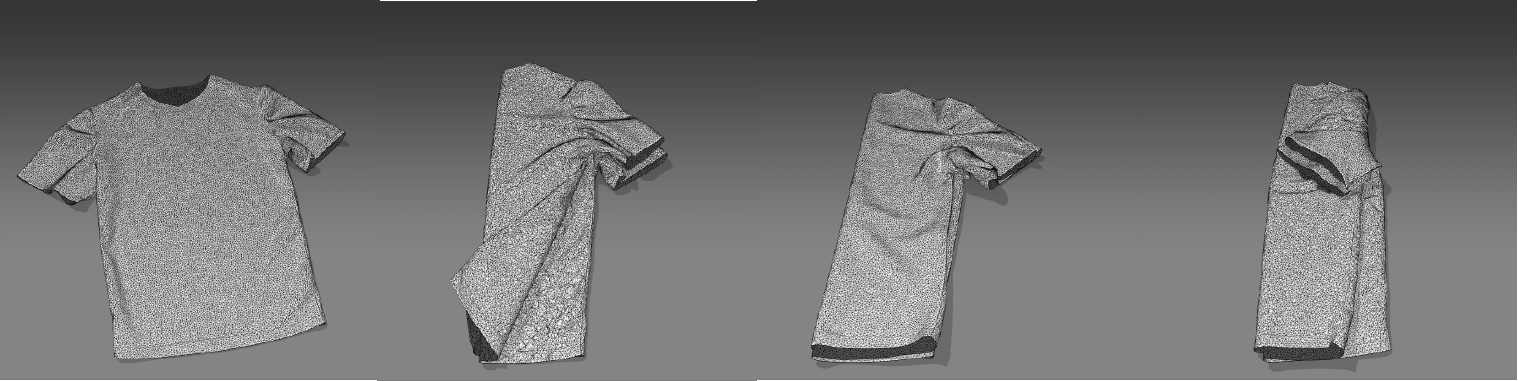}

\caption{\textbf{Folded T-shirt Meshes:} Shown above is the fold sequence the robot is expected to perform during task execution. To simulate the distribution of shirt configurations encountered during folding, we source meshes of t-shirts created by artists and available for purchase online. We standardize the mesh dimensions in Blender, randomize their aspect ratios, and label vertices corresponding to keypoints.}
\label{fig:shirt_folds}
\end{figure}
\subsubsection{OBB Prediction}\label{app:obb_eval}
Similar to the 9DOF pose estimation~\cite{wang2019normalized}, we use 3D intersection over union or (3D IOU) to measure box fit. However, since the boxes computed from OBBs can rotate freely around symmetric axes, we purposely only consider a low acceptance criteria of $ > 0.25$ IOU. In practice, we notice there is significantly high correlation between this metric and grasp success during physical trials. The confidence value used during mAP calculation was the probability density of the predicted Gaussian peak around the object centroid. 

\subsubsection{Keypoint Prediction}\label{app:kp_eval}
To classify whether a ground-truth keypoint was successfully predicted, we check whether a predicted keypoint for the class is within 20 pixels of the ground-truth keypoint. If no such predicted keypoint exists, then we consider it a false negative. If a predicted keypoint is not within 20 pixels of a ground-truth keypoint of the same class, it is considered a false positive. We average over the individual class scores to obtain mAP scores, and each point on each class's PR curve is computed by linearly varying the threshold for detection in the heatmap from 0 to 1.

\subsection{Manipulation Details}\label{app:manip}
In this section, we will describe how robot manipulation policies are constructed for the object grasping and t-shirt folding experiments.

\subsubsection{Novel Object Grasping}\label{app:novel-object-grasping}
Inspired by~\cite{balasubramanian2012physical}, given an OBB,  the robot aligns the gripper with the largest principal axis. Thus, a bottle standing up  would have a "side" grasp where the robot grasps perpendicular to the face of the bottle. However, a stapler would have a "top" grasp where the robot aligns the gripper with the orientation of the object with a vertical approach direction. If an object has no dominant principle axes, like a cube, the robot favors "side" grasps, which require less motion for the kinematics of the HSR. We note this is not meant to be an optimal grasping strategy for all objects, but is only meant to handle a wide variety of common household objects shown during the grasping trials. 

\subsubsection{T-Shirt Folding}\label{app:t-shirt-manip}
Given the 2D key-points, which predict neck, sleeves and bottom corners. We can encode a set of of grasps and pull behaviors that are relative to predicted key-points. Our t-shirt fold sequence is known colloquially as the Sideways Column method. The fold sequence consists of the following steps: 1) pull the left-most sleeve to the right-most sleeve, 2) pull the left-most bottom to the right-most bottom, 3) pull the sleeves inwards to be aligned with the neck of the shift, 4) pull the bottom of the t-shirt onto the top. An illustration of this fold sequence can be seen in Fig. ~\ref{fig:shirt_folds}.

In order to compute the 3D locations of these grasp points, we need to project the 2D key-points onto the scene. We did this by leveraging the predicted disparity information and segmentation of the table. We first fit a 3D plane to the table using the disparity and segmentation output heads. We then project the 2D key-points onto the plane, by computing the ray-to-plane intersection point. Given the 3D locations of the keypoint the robot can then infer grasp position and pull locations.

\section{Perception Ablation Studies}
\label{app:ablation}
We ablate different methods of using different sensing modalities (monocular, depth, RGB-D, stereo) in this section. We describe each baseline implementation in Section~\ref{app:baselines}. Models are trained on the simulated small objects dataset generated with the Basler stereo pair camera model (Section~\ref{app:basler}). We collect a dataset of real images using the Basler stereo pair and annotate them with 3D oriented bounding boxes. This dataset only consists of optically easy scenarios, non-transparent objects with low amounts of natural lighting, and we therefore expect the depth-based networks to perform well. We hope that SimNet is able to roughly match or outperform as well on this task.

We observe that predicting coarse disparity for the monocular, depth, and RGB-D networks results in very little difference in performance. We also find that simply stacking the RGB and D channels performs roughly the same as feeding them into separate feature extractors. Pure monocular reasoning thought does quite bad, but this is likely due to the 3D nature of the problem and not reflective of sim-to-real transfer. SimNet is able to have comparable performance to RGB-D transfer techniques, which suggest similar sim-to-real performance on optically easy scenarios between the methods.

\begin{table}[!htbp]
\centering
% \vspace{-0.2cm}
% \resizebox{\columnwidth}{!}{
% density, knot, rand success, rand actions, hulk success, hulk actions, hulk failures
 \begin{tabular}{| c | c |}
\hline
Method & 3D mAP \\ 
\hline
Mono & 0.164\\
Mono-Aux & 0.169\\
Depth & 0.831\\
Depth-Aux & 0.838\\
RGB-D & 0.855\\
RGB-D-Aux & 0.864\\
RGB-D-Stack & 0.856\\
RGB-D-Seq & 0.774\\
\net{} & \textbf{0.921}\\
\hline
\end{tabular}
% }

\caption{\textbf{Small Object Ablation Study:} We perform an ablation study of different perception procedures and modalities on the small objects dataset. We evaluate each model on real, annotated images on the OBB prediction task. We find that monocular networks perform very poorly, while depth and RGB-D perform much better. However, \net{} consistently performs the well on this dataset of optically easy scenarios.}
\label{table:panoptic_ablation}

\end{table}

\section{Dataset Details}
\label{app:dataset}

\begin{figure}[t!]
    \centering
    \includegraphics[width=5.5in]{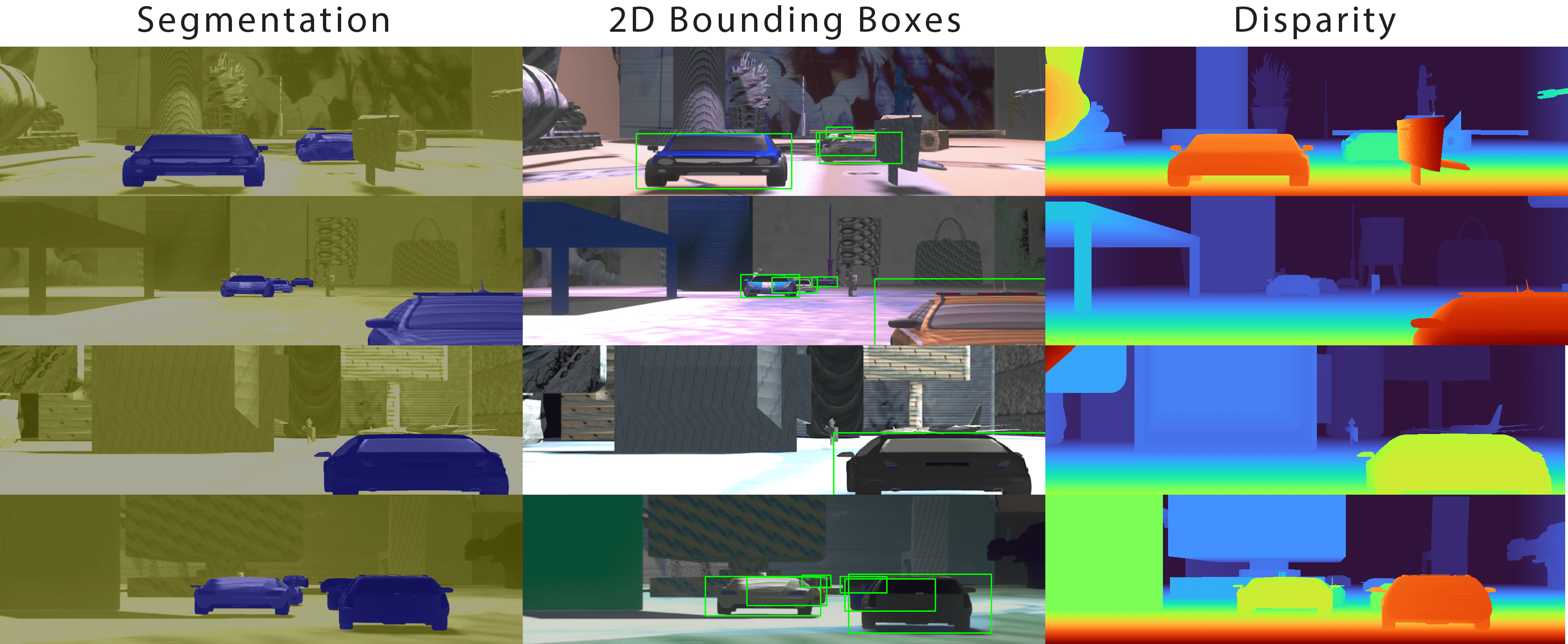}

\caption{\textbf{SimNet Car Dataset:} We simulate cars and large objects in SimNet to train networks for the KITTI object detection benchmark. Each stereo image is labeled in simulation with segmentation masks, bounding boxes, and disparity. We use bounding boxes instead of oriented bounding boxes in this dataset, because the KITTI task requires 2D bounding boxes.}
\label{fig:car_simnet}
\end{figure}

\paragraph{2D Car Detection Dataset:}
To predict 2D bounding boxes for cars on the road, we designed a dataset that enables our network to learn relevant geometric features for cars. Given car poses from held out KITTI~\cite{Geiger2012CVPR,Geiger2013IJRR} scenes, we sampled cars and camera positions to be in natural configurations to the natural world. Our car assets were taken from ShapeNet~\cite{chang2015shapenet}. For distractor objects, we used random meshes from Shapenet and scaled them to be the size of buildings, buses, pedestrians and buses. We then placed them randomly in the scene to be collision free and resting on the ground plane. Images of our generated environments can be seen in Fig. ~\ref{fig:simulator}. The car dataset generated in SimNet contains 50,000 stereo RGB images with annotations for 2D bounding boxes, segmentation masks, and disparity (Figure~\ref{fig:car_simnet}).

\paragraph{Small Objects Dataset:}
For grasping objects on a table, we simulate indoor scenes by having a flat table with random objects from Shapenet~\cite{chang2015shapenet} placed on top of it. Our table is then surrounded by other random furniture from Shapenet that is randomly placed juxtaposed to the table. We then vary the size of the room and camera camera position relative to the table. Specifically, we sampled randomly positions from a half-sphere with a radius that is uniformly sampled between $[0.5,2]$ meters. 

\begin{figure}[t!]
    \centering
    \includegraphics[width=5.5in]{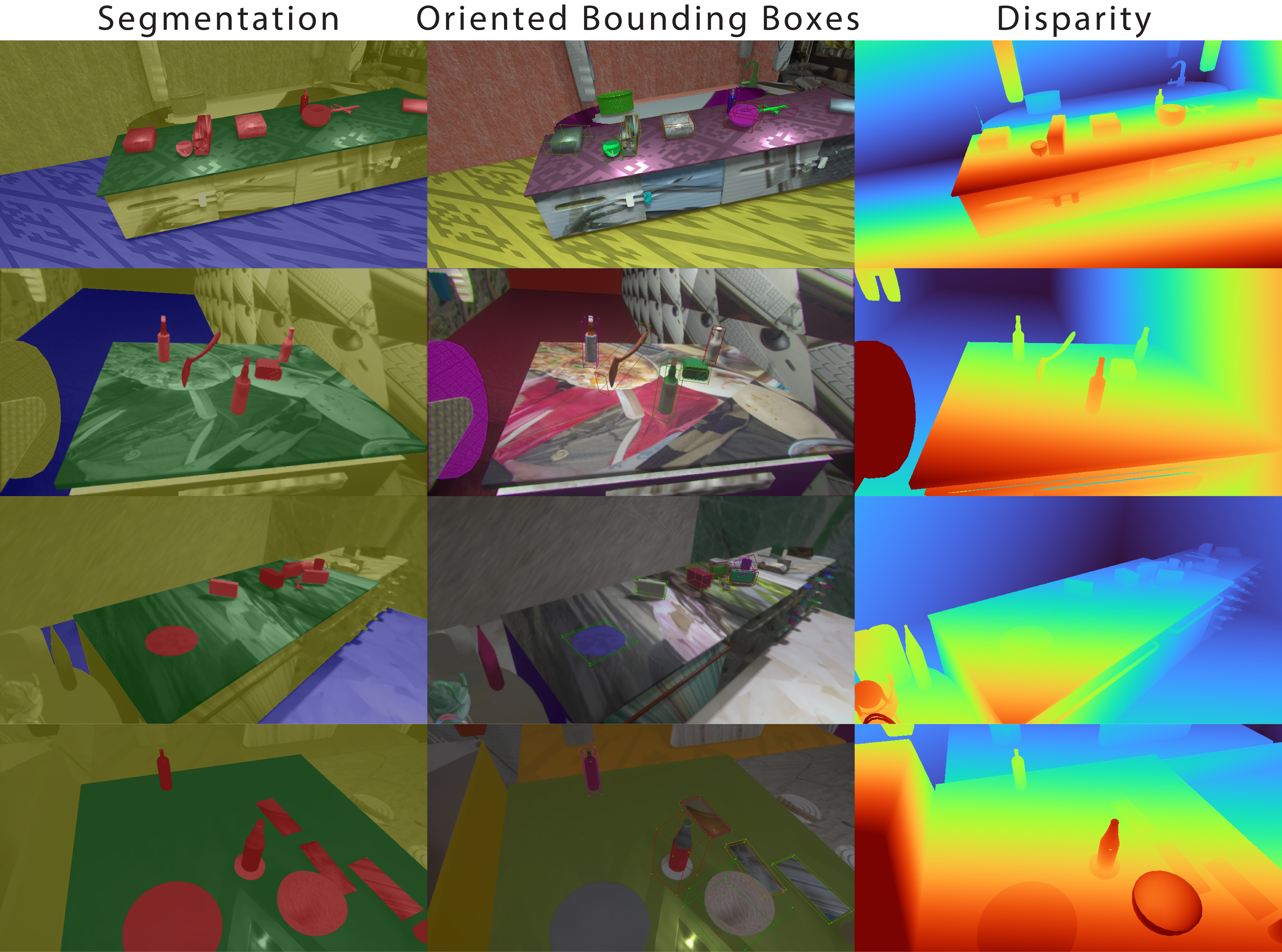}

\caption{\textbf{SimNet Small Objects Dataset:} We simulate small objects placed on tabletops in SimNet. The dataset consists of stereo images with annotations for segmentation masks, oriented bounding boxes, and disparity.}
\label{fig:grasp_simnet}
\end{figure}

This dataset consists of 50,000 stereo RGB images generated in SimNet.  Each stereo image pair contains full annotations for full-resolution disparity images, OBBs, class and segmentation masks, as shown in Fig.  \ref{fig:grasp_simnet}.The same training set (and networks) is used for evaluation on both types of real objects (easy and hard). To train the RGB-D models, we make an equivalent dataset of simulated RGB-D images instead. We add structured noise to depth images to simulate artifacts commonly found in real images (Figure~\ref{fig:depth_noise}). For the real grasping experiments, we use the Zed 2 stereo pair for \net{} training and prediction and the HSR's Asus Xtion for RGB-D. For perception ablation studies, we use the Basler stereo pair and Kinect camera models, which are described in Section~\ref{app:basler}. 

\paragraph{T-shirt Dataset:}
We compile a dataset of 50 meshes of t-shirts and polos in various stages of folding that were purchased online (Figure~\ref{fig:shirt_folds}). We randomize the scale and aspect ratios of each shirt mesh before placing it on a flat table in the scene with surrounding furniture as in the small objects dataset. Each scene has 1-3 t-shirts, and we generate keypoint, segmentation, OBB, and full-resolution disparity annotations in simulation. This dataset consists of 50,000 stereo RGB images. Each stereo image pair contains annotations for full-resolution disparity images projected onto the left image, segmentation masks, oriented bounding boxes, and keypoints. 

\begin{figure}[t!]
    \centering
    \includegraphics[width=5.5in]{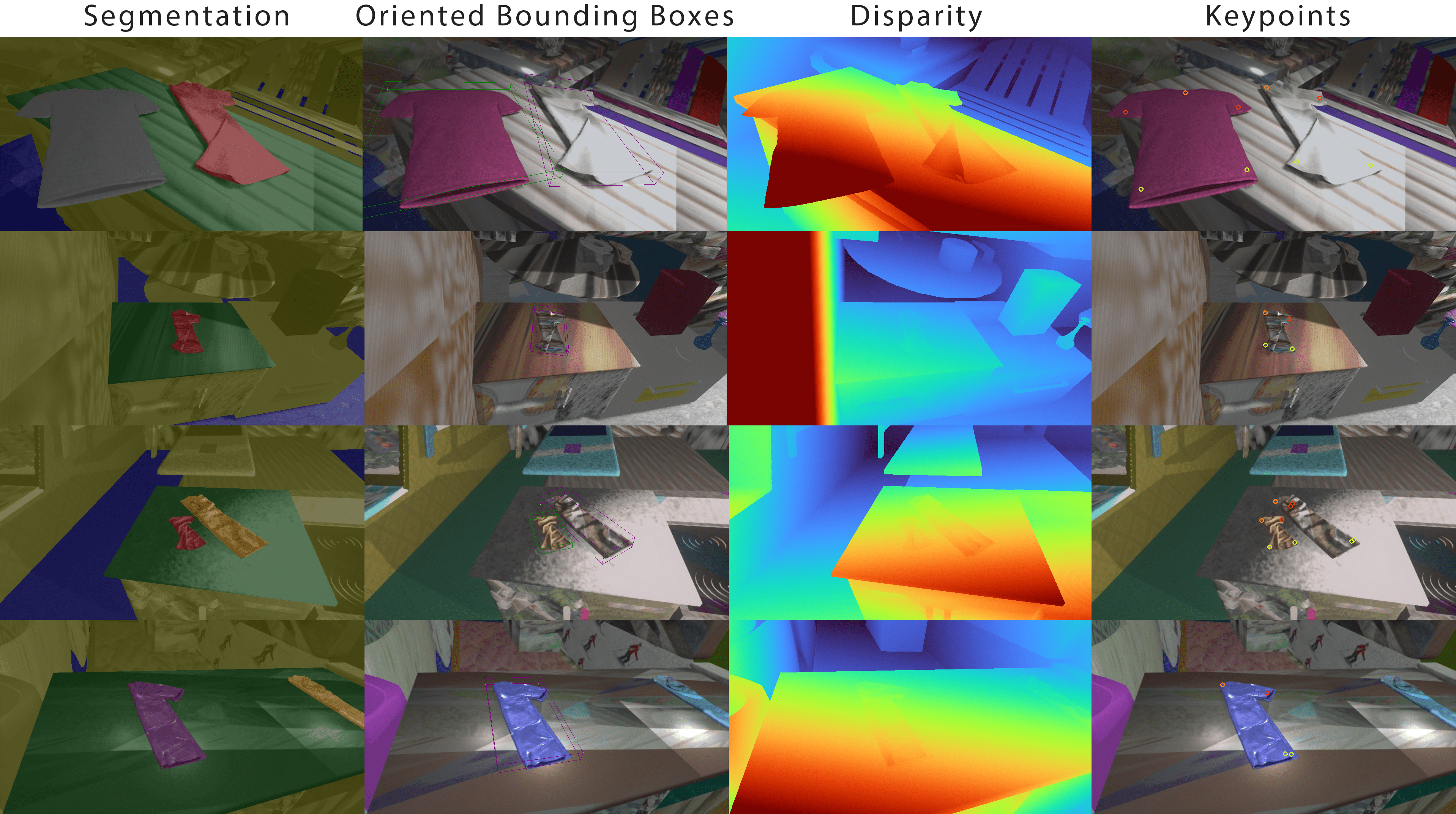}

\caption{\textbf{SimNet T-Shirt Folding Dataset:} We simulate t-shirts in various stages of folding on flat surfaces in the center of the workspace. We procedurally annotate each stereo image with segmentation masks, oriented bounding boxes, depth images, and keypoints. The keypoints correspond to the sleeves, neck, and bottom corners of each shirt. In this dataset, we label each stage of folding with different segmentation classes. This is useful for grasp failure detection when folding with the robot.}
\label{fig:shirt_simnet}
\end{figure}

The validation dataset is collected on the Toyota HSR using a mounted Zed 2 stereo pair and consists of 32 images. Each real image is manually annotated with keypoints. We consider shirt folding as a sequence of $4$ folds (Figure~\ref{fig:shirt_folds}), and we label each stage of folding with a different semantic segmentation class. This is not strictly neccessary, but useful for failure detection when conducting robot experiments. If a grasp is missed, which is possible due to the large morphology of the gripper, this enables the planner to recognize that it is still in the same state.  For the depth baselines, we use the Asus Xtion camera on the Toyota HSR. We present visualizations in Figure~\ref{fig:shirt_simnet}.

% % \subsection{Cars}
% %  The scene consists of car meshes and large objects randomly placed in the scene. \brijen{describe more details}
% \subsection{Panoptic Objects}
%  The scene consists of several flat surfaces in the center of a room, with up to 15 objects lying on each surface. 
% \subsection{Shirts}
% 